\documentclass[10pt,twocolumn,letterpaper]{article}

\usepackage[pagenumbers]{cvpr} 

\usepackage[dvipsnames]{xcolor}

\definecolor{cvprblue}{rgb}{0.21,0.49,0.74}

\usepackage{array}
\usepackage{subcaption}
\usepackage{multirow,multicol}
\usepackage[flushleft]{threeparttable}
\usepackage{comment}
\usepackage{algorithmic}
\usepackage{booktabs}
\usepackage{bbm, dsfont}
\usepackage[font=small,labelfont=bf]{caption}

\usepackage{amssymb}
\usepackage{pifont}

\newcommand{\cmark}{\text{\ding{51}}}%
\newcommand{\xmark}{\text{\ding{55}}}%

\newcommand{\tablestyle}[2]{\setlength{\tabcolsep}{#1}\renewcommand{\arraystretch}{#2}\centering\footnotesize}

\newcolumntype{x}[1]{>{\centering\arraybackslash}p{#1pt}}
\newcommand{\app}{\raise.17ex\hbox{$\scriptstyle\sim$}}

\newlength\savewidth\newcommand\shline{\noalign{\global\savewidth\arrayrulewidth
  \global\arrayrulewidth 1pt}\hline\noalign{\global\arrayrulewidth\savewidth}}

\makeatletter\renewcommand\paragraph{\@startsection{paragraph}{4}{\z@}
  {.5em \@plus1ex \@minus.2ex}{-.5em}{\normalfont\normalsize\bfseries}}\makeatother

\def\tablecite#1#{%
  \def\pretablecite{#1}%
  \tableciteaux}
\def\tableciteaux#1{%
  \textsuperscript{\expandafter\originalcite\pretablecite{#1}}%
}

\usepackage{graphicx}
\usepackage{enumitem}
\usepackage{wrapfig}
\usepackage{lipsum}
\usepackage{soul}

\usepackage{color, colortbl}

\usepackage{tabu}
\usepackage{xcolor}
\usepackage{nicematrix}

\definecolor{ForestGreen}{rgb}{0.13, 0.55, 0.13}
\definecolor{Green}{rgb}{0.0, 0.5, 0.0}
\definecolor{Blue}{rgb}{0.25, 0.42, 0.88}
\definecolor{green(munsell)}{rgb}{0.0, 0.66, 0.47}
\definecolor{green(ryb)}{rgb}{0.4, 0.69, 0.2}
\definecolor{green(pigment)}{rgb}{0.0, 0.65, 0.31}
\definecolor{citecolor}{HTML}{0071bc}
\definecolor{GrayXMark}{gray}{0.7}
\definecolor{DifferenceColor}{HTML}{af3235}
\definecolor{HighlightColor}{gray}{0.9}
\definecolor{OracleTextColor}{gray}{0.55}
\definecolor{Cerulean}{HTML}{00a2e3}

\newcommand{\rownumber}[1]{\textcolor{Cerulean}{#1}}
\newcolumntype{H}{>{\setbox0=\hbox\bgroup}c<{\egroup}@{}}
\newcolumntype{a}{>{\columncolor{HighlightColor}}c}
\newcolumntype{L}[1]{>{\centering\arraybackslash}m{#1}}
\DeclareRobustCommand{\colorrowtext}[0]{{\sethlcolor{HighlightColor}\hl{gray}}}
\DeclareRobustCommand{\bgcolortext}[1]{{\sethlcolor{HighlightColor}\hl{#1}}}

\usepackage{tabularx}
\usepackage[export]{adjustbox}
\usepackage{makecell}
\usepackage{ragged2e}

\newcommand{\plus}[1]{\small\bf\textcolor{Green}{#1}}
\newcommand{\diff}[1]{\scriptsize\bf\textcolor{DifferenceColor}{(#1)}}

\newcommand{\ours}{InstanceDiffusion\xspace}
\newcommand{\unifusion}{UniFusion\xspace}
\newcommand{\maskedattn}{Instance-Masked Attention\xspace}
\newcommand{\scaleu}{ScaleU\xspace}
\newcommand{\mis}{Multi-instance Sampler\xspace}
\newcommand{\textToI}{text-to-image\xspace}

\newcommand{\Scribble}{Scribble\xspace}
\newcommand{\scribble}{scribble\xspace}
\newcommand{\ccolor}{\cellcolor{gray!20}}
\newcommand{\coco}{COCO\xspace}

\newcommand{\model}{f}
\newcommand{\bc}{\mathbf{c}}
\newcommand{\bl}{\mathbf{l}}
\newcommand{\bp}{\mathbf{p}}
\newcommand{\bg}{\mathbf{g}}
\newcommand{\bG}{\mathbf{G}}
\newcommand{\bv}{\mathbf{v}}
\newcommand{\bV}{\mathbf{V}}
\newcommand{\bM}{\mathbf{M}}
\newcommand{\bF}{\mathbf{F}}
\newcommand{\bs}{\mathbf{s}}
\newcommand{\balpha}{\boldsymbol{\alpha}}

\newcommand{\gxmark}{\textcolor{gray}{\xmark}}
\usepackage{dblfloatfix}

\usepackage[pagebackref,breaklinks,colorlinks,citecolor=cvprblue]{hyperref}

\usepackage[capitalize]{cleveref}
\crefname{section}{\S}{\S\S}
\crefname{subsection}{\S}{\S\S}
\crefformat{table}{Table~#2#1#3}
\crefformat{figure}{Figure~#2#1#3}
\crefformat{equation}{Eq~#2#1#3}

\def\paperID{240} 
\def\confName{CVPR}
\def\confYear{2024}

\title{\ours: Instance-level Control for Image Generation}

\author{
  Xudong Wang$^{1,2}$ \quad 
  Trevor Darrell$^{2}$ \quad 
  Sai Saketh Rambhatla$^{1}$ \quad 
  Rohit Girdhar$^{1}$ \quad 
  Ishan Misra$^{1}$ \quad 
   \\ \vspace{4pt}
   $^{1}$GenAI, Meta\quad \quad $^{2}$UC Berkeley 
   \\
   \small{project page:} \href{https://people.eecs.berkeley.edu/~xdwang/projects/InstDiff/}{\small{https://people.eecs.berkeley.edu/$\sim$xdwang/projects/InstDiff/}}
}

\begin{document}






\twocolumn[{%
  \renewcommand\twocolumn[1][]{#1}%
  \maketitle
    \vspace{-12pt}
    \captionsetup{type=figure}
    \centering
    \includegraphics[width=0.99\textwidth]{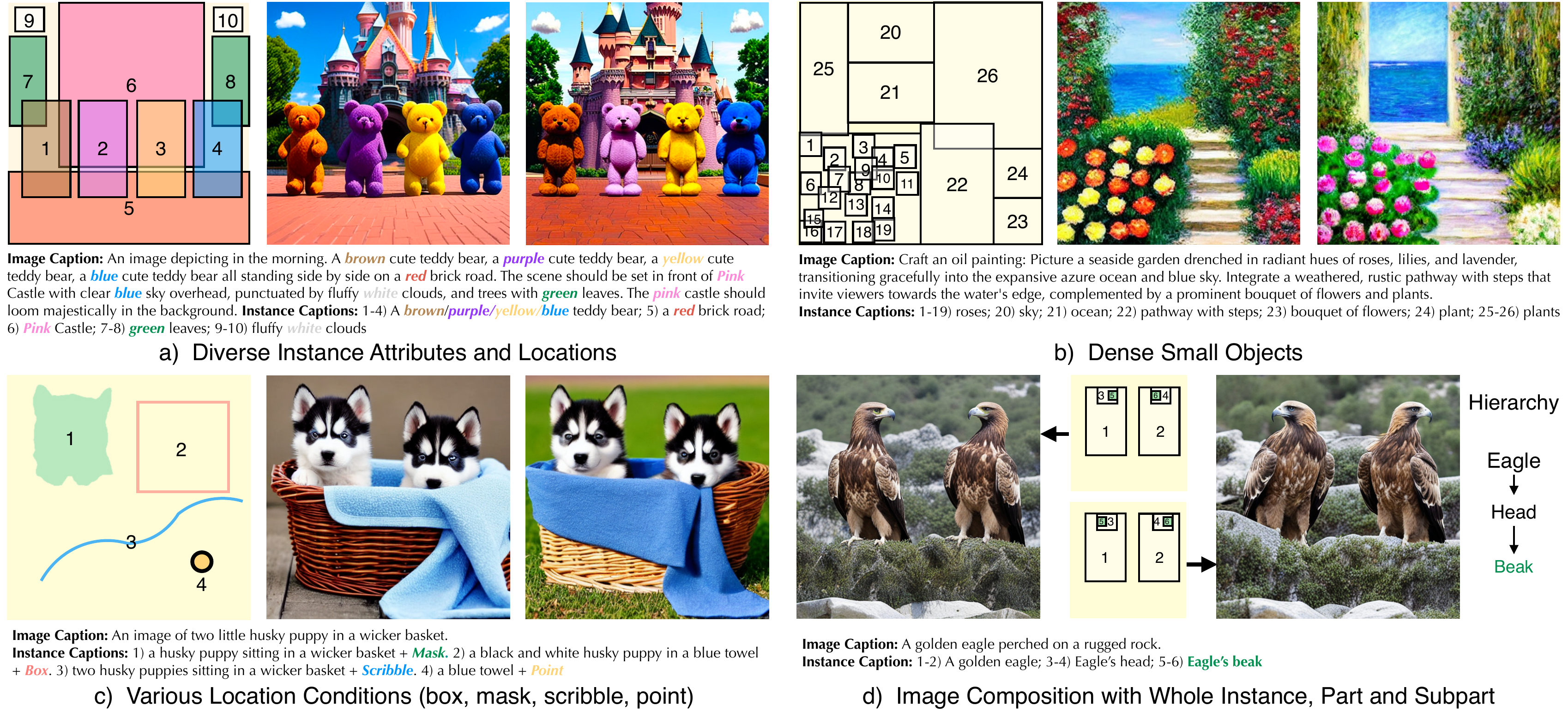}
    \vspace{-4pt}
    \caption{\ours's generations using instance-level text prompts and location conditions for image generation. Our model can respect: a) a variety of instances with diverse attributes (8 colors) and boxes, b) densely-packed instances ($>$25 objects), c) mixed location conditions (such as boxes, masks, \scribble{}s, and points), and d) compositions with granularity spanning from entire instances to parts and subparts. 
    The positioning of parts/subparts implicitly alters the overall pose of the object.
    The instance inputs and their global text prompts are displayed, with the location conditions displayed on the left image. 
    Numbers in the box/mask/scribble/point refer to the instance id. 
    }
    \label{fig:teaser}
    \vspace{18pt}
}]

\begin{abstract}
Text-to-image diffusion models produce high quality images but do not offer control over individual instances in the image.
We introduce \ours that adds precise instance-level control to text-to-image diffusion models.
\ours supports free-form language conditions per instance and allows flexible ways to specify instance locations such as simple single points, \scribble{s}, bounding boxes or intricate instance segmentation masks, and combinations thereof.
We propose three major changes to text-to-image models that enable precise instance-level control.
Our UniFusion block enables instance-level conditions for text-to-image models, the ScaleU block improves image fidelity, and our Multi-instance Sampler improves generations for multiple instances.
\ours significantly surpasses specialized state-of-the-art models for each location condition.
Notably, on the COCO dataset, we outperform previous state-of-the-art by 20.4\% AP$_{50}^\text{box}$ for box inputs, and 25.4\% IoU for mask inputs.
\end{abstract}

\section{Introduction}
\label{sec:intro}

Image generation models~\cite{reed2016generative,goodfellow2020generative,karras2017progressive,brock2018large,karras2020analyzing,zhu2017unpaired,ho2020denoising,song2020denoising,saharia2022photorealistic,ramesh2022hierarchical,chang2023muse} trained on web-scale data have made tremendous progress in the recent years.
Notably, text conditioned diffusion models now produce high quality images that contain the free form concepts specified in the text~\cite{ho2020denoising,song2020denoising,dhariwal2021diffusion,song2020score,saharia2022photorealistic,ramesh2022hierarchical}.
While text-based control is useful, it does not always allow for precise and intuitive control over the output image.
Thus, many different forms of conditioning, \eg, edges, normal maps, semantic layouts have been proposed for better control~\cite{nichol2021glide,li2023gligen,zhang2023adding,gafni2022make,feng2023layoutgpt,goel2023pair,park2022shape,zhang2020text,bau2021paint,avrahami2022blended}.
These richer controls enable a broader range of applications for the generative models in design, data generation~\cite{ge2022dall,Zhao2022XPasteRC} \etc.
In this work, we focus on precise control over the \textit{instances} in terms of their location and attributes in the output image.

We propose and study instance-conditioned image generation whereby a user can specify \emph{every} instance in terms of its location and an instance-level text prompt to generate an image.
The location can be specified using either a bounding box, an instance mask, a single point or a \scribble.
Practically, this allows for a flexible input where some instance locations maybe specified more precisely using masks, and others less precisely using points.
The per instance text prompts allow for fine-grained control over the instance's attributes such as color, texture, \etc.
Our proposed instance-conditioned generation is a generalization of settings studied in prior work~\cite{avrahami2023spatext,li2023gligen,zhang2023adding} that consider only one location format and do not use per instance captions.

Our model presents several design choices that enable more precise yet flexible control for instances in the output image.
Since locations can be specified in a variety of formats, we present a unified way to parameterize and fuse their information during the generation process. 
Our unified modeling is simpler than prior work that uses separate architectures and strategies to model different location formats.
Moreover, the unified modeling of location formats allows the model to exploit the shared underlying structure of instance locations which improves performance.

Through comprehensive evaluations, our method \ours outperforms state-of-the-art models specialized for particular instance conditions.
We achieve a 20.4\% increase in AP$_{50}^\text{box}$ over GLIGEN~\cite{li2023gligen} when evaluating with bounding box inputs on COCO~\cite{lin2014microsoft} \texttt{val}.
For mask-based inputs, we obtain a {25.4}\% boost in IoU compared to DenseDiffusion~\cite{densediffusion} and a {36.2}\% gain in AP$_{50}^\text{mask}$ over ControlNet~\cite{zhang2023adding}.
As prior methods do not study point or \scribble inputs for image generation, we introduce evaluation metrics for these settings.
\ours also demonstrates superior ability to adhere to attributes specified by instance-level text prompts.
We obtain a substantial 25.2 point gain in instance color accuracy and a 9.2 point improvement in texture accuracy compared to GLIGEN.

\par \noindent \textbf{Contributions.}
(1) In this paper, we propose and study instance-conditioned image generation that allows flexible location and attribute specification for multiple instances.
(2) We propose three key modeling choices that improve results --
(i) \textit{{UniFusion}} (\cref{sec:method-unifusion}), which projects various forms of instance-level conditions into the same feature space, and injects the instance-level layout and descriptions into the visual tokens;
(ii) \textit{{ScaleU}} (\cref{sec:method-scaleu}), which re-calibrates the main features and the low-frequency components within the skip connection features of UNet, enhancing the model's ability to precisely adhere to the specified layout conditions;
(iii) \textit{{Multi-instance Sampler}} (\cref{sec:method-mis}), which reduces information leakage and confusion between the conditions on multiple instances (text+layout).
(3) A dataset with instance-level captions generated using pretrained models (\cref{sec:method-data}) and a new set of evaluation benchmarks and metrics for measuring the performance of location grounded image generation (\cref{sec:exp-metrics}).
(4) Our unified modeling of different location formats significantly improves results over prior work (\cref{sec:exp-results}).
We also show that our findings can be applied to previous approaches and boost their performance.

\section{Related Work}
\label{sec:related}
\noindent \textbf{Image Diffusion Models}~\cite{sohl2015deep,ho2020denoising,song2020score} learn the process of text-to-image generation through iterative denoising steps initiated from an initial random noise map.
Latent diffusion models (LDMs)~\cite{rombach2021highresolution,van2017neural} perform the diffusion process in the latent space of a Variational AutoEncoder~\cite{DBLP:journals/corr/KingmaW13,van2017neural}, for computational efficiency, and encode the textual inputs as feature vectors from pretrained language models~\cite{radford2021learning}.
DALL-E 2~\cite{ramesh2022hierarchical} synthesizes images using the image space of CLIP~\cite{radford2021learning}. 
In contrast, Imagen~\cite{saharia2022photorealistic} diffuses pixels directly, without the need for latent images. In addition, it demonstrates that generic large language models, such as T5~\cite{raffel2020exploring}, trained solely on text corpora, are surprisingly effective at encoding text for image generation.

\noindent \textbf{Image Generation with Spatial Controls} is a form of conditional image synthesis task~\cite{zhu2017unpaired,liu2017unsupervised,isola2017image,wang2018video,wang2018high,xu2018attngan,zhang2021ufc,van2016conditional,hertz2022prompt,gafni2022make,li2023gligen,zhang2023adding,feng2023layoutgpt,stap2020conditional}, which introduces spatial conditioning controls to guide the image generation process.
\noindent \textit{Make-a-Scene}, \textit{SpaText}~\cite{avrahami2023spatext}, \textit{GLIGEN}~\cite{li2023gligen}, and \textit{ControlNet}~\cite{zhang2023adding} add finer grained spatial control, such as semantic segmentation masks, to large pretrained diffusion models by allowing users to include additional images that explicitly define their desired image composition.
\textit{GLIGEN}~\cite{li2023gligen} can also support controlled image generation using discrete conditions such as bounding boxes.
MultiDiffusion~\cite{bar2023multidiffusion}, DenseDiffusion~\cite{densediffusion}, Attend-and-Excite~\cite{chefer2023attend}, ReCo~\cite{yang2023reco}, StructureDiffusion~\cite{feng2022training}, Layout-Guidance~\cite{chen2023trainingfree}, and BoxDiff~\cite{xie2023boxdiff} add location controls to diffusion models without fine-tuning the pretrained text-to-image models. 
\noindent \textbf{Discussions.} 
ControlNet and GLIGEN require training separate models for each type of controllable input, which increases the overall complexity of the system and not effectively capture interactions across various controllable inputs. 
Moreover, while ControlNet focuses solely on spatial conditions and GLIGEN employs \textit{object category} as the text prompt, the lack of training the models with detailed instance-level prompts not only limits user control but also hinders the model from effectively leveraging instance descriptions. 

\def\figInstDiff#1{
    \captionsetup[sub]{font=small}
    \begin{figure}[#1]
      \centering
      \includegraphics[width=0.99\linewidth]{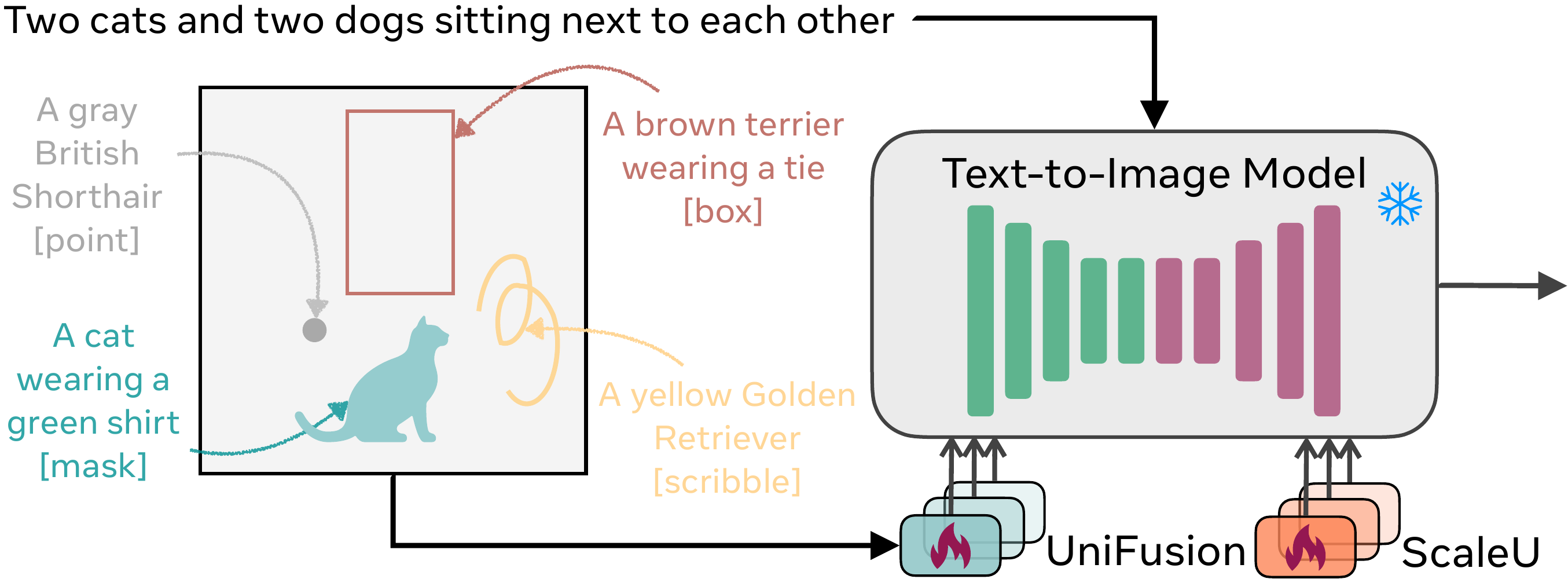}\vspace{-3pt}
      \caption{\textbf{\ours} enhances \textToI models by providing additional instance-level control.
      In additon to a global text prompt, \ours allows for paired instance-level prompts and their locations to be specified when generating images.
      \ours is versatile, supporting a range of location forms, from the simplest points, boxes, and \scribble{}s to more complex masks, and their flexible combinations.
      }
      \label{fig:instDiff}
    \end{figure}
}

\def\figMIS#1{
    \captionsetup[sub]{font=small}
    \begin{figure}[#1]
    \centering
    \includegraphics[width=0.99\linewidth]{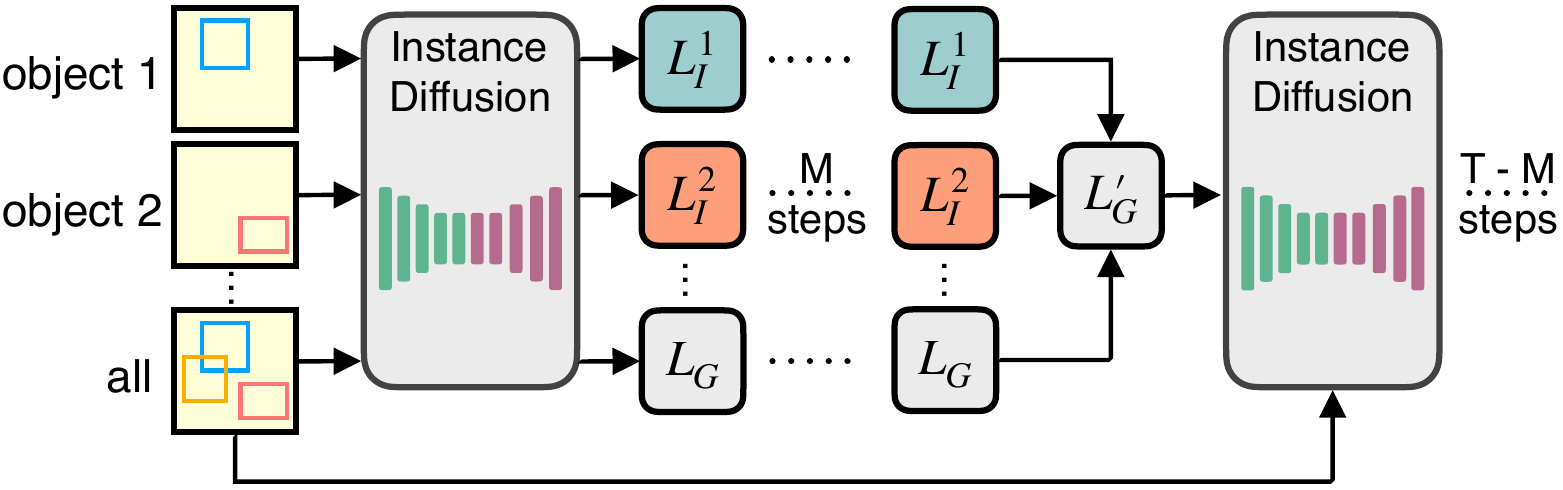}\vspace{-5pt}
    \caption{Model inference with \textbf{\mis} to minimize information leakage across multiple instance conditionings.
    }
    \label{fig:mis}
    \end{figure}
}

\def\figUniFusion#1{
    \captionsetup[sub]{font=small}
    \begin{figure}[#1]
      \centering
      \includegraphics[width=1.0\linewidth]{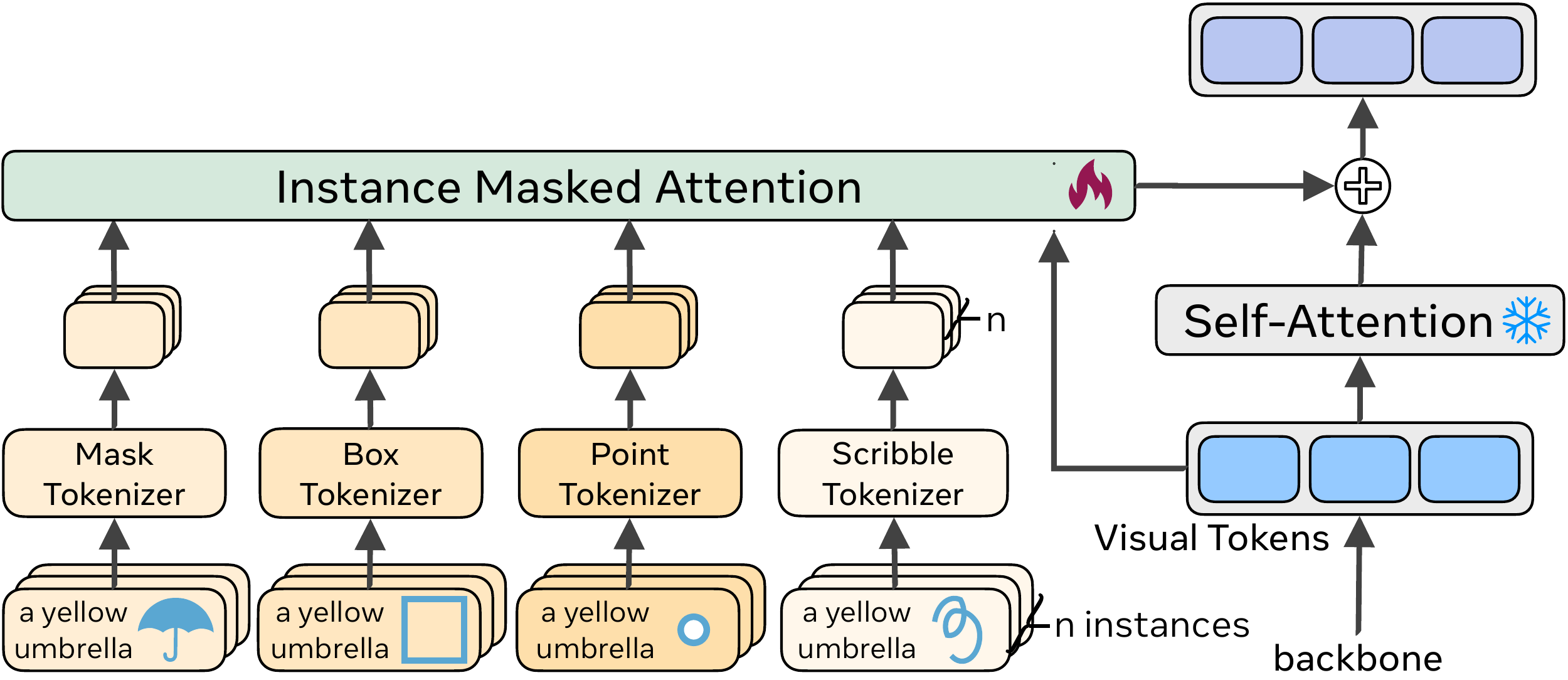}\vspace{-3pt}
      \caption{\textbf{\unifusion} projects various forms of instance-level conditions into the same feature space, seamlessly incorporating instance-level locations and text-prompts into the visual tokens from the diffusion backbone.}
      \label{fig:unifusion}
    \end{figure}
}

\def\figInputs#1{
    \captionsetup[sub]{font=small}
    \begin{figure}[#1]
      \centering
      \includegraphics[width=1.0\linewidth]{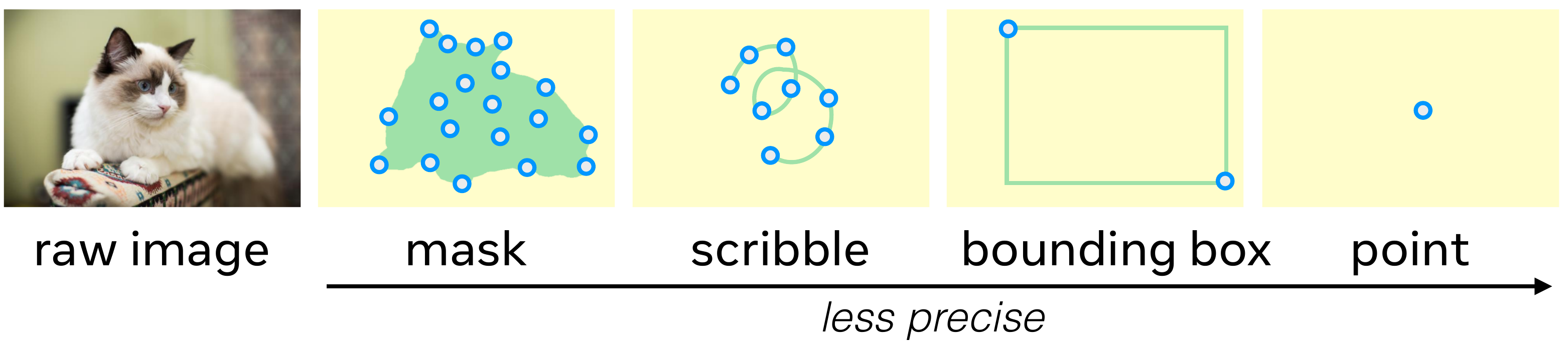}\vspace{-8pt}
      \caption{We represent different location condition formats as sets of \textcolor{Blue}{points}, with each format having varying quantities of points. Masks are represented as sparsely sampled points within the mask and uniformly sampled points from boundary polygons, bounding boxes by the top-right and bottom-right corners, and \scribble are converted into uniformly sampled points.}
      \label{fig:input}
    \end{figure}
}


\section{Instance Diffusion}
\label{sec:method}
\figInstDiff{t!}
We study adding precise, versatile instance-level control for text-based image generation.

\par \noindent \textbf{Problem definition.}
We aim to improve instance-level control in image generation by focusing on two conditioning inputs for each instance, namely, its location and a text caption describing the instance.
More formally, we want to learn an image generation model $\model(\bc_g, \{(\bc_1, \bl_1), \ldots, (\bc_n, \bl_n)\})$ that is conditioned on a global text caption $\bc_g$ and the per-instance conditions $(\bc_i, \bl_i)$ containing
caption $\bc_i$ and location $\bl_i$ for $n$ instances. 
This problem is similar to~\cite{avrahami2023spatext} and is a generalization of the `open-set grounded text-to-image'~\cite{li2023gligen} problem which does not consider per-instance captions.
Our generalization allows for a generic and flexible way to control the scene-layout in terms of locations and attributes of the instances, as well as scene-level control via the global caption.

\subsection{Approach overview}
We introduce \ours (\cref{fig:instDiff}) for instance-conditioned image generation using a diffusion model.
We consider a variety of different and flexible ways to specify an object's location, \eg, a single point, a \scribble, a bounding box, and an instance mask.
Since obtaining large-scale paired (text, image) data is much easier compared to (instance, image) data, we use a pretrained \textToI UNet model that is kept frozen.
We add our proposed learnable \textbf{\unifusion} blocks to handle the additional per-instance conditioning. \unifusion fuses the instance conditioning with the backbone and modulate its features to enable instance-conditioned image generation.
Additionally, we propose \textbf{\scaleu} blocks that improve the UNet's ability to respect instance-conditioning by rescaling the skip-connection and backbone feature maps produced in the UNet.
At inference, we propose \textbf{\mis} which reduces information leakage across multiple instances.

Since obtaining a large paired (instance, image) dataset is difficult, we automatically generate a dataset with instance-level location and text captions using state-of-the-art recognition systems.
Finally, we propose a new and comprehensive benchmark to evaluate the model's performance for instance-conditioned generation.

\figUniFusion{t!}
\figInputs{t!}
\subsection{\unifusion block}
\label{sec:method-unifusion}

The \unifusion block, illustrated in~\cref{fig:unifusion}, tokenizes the per-instance conditions $(\bc_i, \bl_i)$ and fuses them with the features, \ie, visual tokens from the frozen \textToI{} model.
Similar to~\cite{alayrac2022flamingo,li2023gligen}, the \unifusion block is added between the self-attention and cross-attention layers of the backbone.
The per-instance location $\bl_i$ can be specified in \textit{one or more} location formats such as masks, boxes, \etc.
We now describe the key operations in the \unifusion block.

\par \noindent \textbf{Location parameterization}.
As shown in~\cref{fig:input}, we convert the four location formats - masks, boxes, \scribble{}s, single point - into 2D points (denoted as $\bp_i\!=\!\{(x_k, y_k)\}^n_{k=1}$ for instance $i$), with each `format' having varying quantities of points $n$.
A \scribble is converted into a set of uniformly sampled points along the curve.
We parameterize bounding boxes by their top-left and bottom-right corners. 
For \textit{instance} masks, we convert them into a set of points sampled from within the mask and from boundary polygons.

\noindent \textbf{Instance Tokenizer}.
We convert the 2D point coordinates $\bp_i$ for each location using a Fourier mapping~\cite{tancik2020fourfeat} $\gamma(\cdot)$ and encode the text prompt $\bc_i$ using a CLIP text encoder $\tau_{\theta}(\cdot)$.
Finally, we concatenate the location and text embeddings and feed them to an MLP to obtain a single token embedding $\bg_i$ for the instance $i$: $\bg_i = \text{MLP}([\tau_{\theta}(\mathbf{c}_i),\gamma(\mathbf{p}_i)])$.
We use a different MLP for each location format.
Moreover, the per-instance location $\bl_i$ can be specified in one or more location formats.
Thus, for each instance $i$, we obtain $\bg^{\mathrm{mask}}_i$, $\bg^{\mathrm{\scribble}}_i$, $\bg^{\mathrm{box}}_i$, and $\bg^{\mathrm{point}}_i$.
If an instance location is specified only using one format, \eg, a single point, we
use a learnable null token $\mathbf{e}_i$ for the other location formats:
\begin{equation}
\vspace{-1pt}
  \bg_i = \text{MLP}([\tau_{\theta}(\mathbf{c}_i), s\cdot\gamma(\mathbf{p}_i)+ (1-s)\cdot\mathbf{e}_i])
\vspace{-1pt}
\end{equation}
where $s$ is a binary value indicating the presence of a specific location format.

\noindent (Optional) To better align with instance mask conditions, we can optionally add extra tokens from binary \textit{instance} masks (dimensions N$\times$H$\times$W, with N as the instance number). These masks are resized to 512$\times$512, and ConvNeXt-tiny~\cite{liu2022convnet} is used to create an 16$\times$16 feature map. The feature map is then flattened into grounding tokens and concatenated with \{$\bg^{\mathrm{mask}}_i$\}$_{i=1}^{n}$. 
These additional mask tokens may offer a minor boost in quantitative performance, yet enhance the model's accuracy in respecting object boundaries.

Prior work resizes \textit{semantic} masks~\cite{zhang2023adding,li2023gligen} 
into the diffusion latent space of size $64\times64$, subsequently adding them into UNet inputs as extra channels.
\textit{Instances from the same semantic class are represented by one mask.}
However, we found that this design choice hurts the performance, particularly in cases with overlapping instances and small objects.


\par \noindent \textbf{\maskedattn and Fusion Mechanism}.
We denote the instance condition tokens, $\bg$, per location format for all $n$ instances by $\bG$, and the $m$ visual tokens, $\bv$, from the backbone as $\bV$.
We apply masked self-attention (SA) to the instance condition tokens and the backbone features
\vspace{-3pt}
\begin{equation}
        \widetilde{\bV}=\mathrm{SA}_{\mathrm{mask}}([\bV, \bG^{\mathrm{mask}}, \bG^{\mathrm{\scribble}}, \bG^{\mathrm{box}}, \bG^{\mathrm{point}}])
    \label{eqn:fusion}
    \vspace{-1pt}
\end{equation}

We consider two design choices, ablated in~\cref{tab:abl-components}, for the location inputs in~\cref{eqn:fusion}:
1) `Format aware' (default) described above models each location format independently via concatenation.
2) `Joint format' jointly models all location formats by concatenating embeddings from each format and converting them into a single embedding (via an MLP) to use in the masked self-attention.

We observed that vanilla self-attention, without masking, led to information leakage across instances, \eg, color of one instance bleeding into another.
Thus, we construct a mask $\bM$ that prevents such leakage across instances:
\begin{equation}
\vspace{-3pt}
\begin{split}
  \text{mask for}\; \bv_k\cdot \bv_j^T&: \bM_{k,j}\!=\!-\inf \; \text{if} \; I_{\bv_k} \neq I_{\bv_j} \\
  \text{mask for}\; \bv_k\cdot \bg_i^T&: \bM_{k,m+i}\!=\!-\inf \; \text{if} \; I_{\bv_k} \neq i \\
\end{split}
\vspace{-2pt}
\end{equation}
where $I_{\bv_k}\!=\!i$ if the visual token $\bv_k$ falls within the region of the instance $i$ defined by either a bounding box or an instance segmentation mask.

Finally, the output of the masked self-attention is added back to the backbone via gated addition
\begin{equation}
\vspace{-2pt}
    \bV=\bV + \tanh(\omega) \widetilde{\bV}[:\!m]
\vspace{-1pt}
\end{equation}
where $\omega$ is a learnable parameter, initialized to $0$, that controls the conditioning contribution of \unifusion.

\subsection{ScaleU block}
\label{sec:method-scaleu}
In the UNet model, each block merges the main feature map $\bF_b$ with the lateral skip-connection features $\bF_s$, passing the concatenated feature to the subsequent UNet block.
FreeU~\cite{si2023freeu} finds that the main backbone of UNet is critical for denoising, whereas its skip connections primarily contribute high-frequency features to the decoder. 
Concatenating these two features directly leads to the network neglecting the semantic content of the main features~\cite{si2023freeu}.
Therefore, FreeU suggests reducing the low-frequency components of the skip features and enhancing the main features using \textit{channel-independent} and \textit{empirically-tuned} values.

Our findings, however, demonstrate that for instance-conditioned image generation, a notable improvement can be achieved by using \textit{channel-wise} and \textit{learnable} vectors to dynamically re-calibrate $\bF_b$ and $\bF_s$. 
More specifically, we introduce \scaleu, that has two \textit{learnable}, \textit{channel-wise} scaling vectors: $\bs_b$, $\bs_s$ for the main and skip-connected features, respectively.
The main features $\bF_b$ are scaled by a simple channel-wise multiplication:
    $\boldsymbol{F}_b'\!=\!\bF_b \otimes (\tanh(\bs_b) + 1)$.
For the skip-connection features, we select the low-frequency (less than $r_{\text{thresh}}$) components using a frequency mask $\balpha$ and scale them in the Fourier domain:
    $\bF'_s\!=\!\text{IFFT}(\text{FFT}(\bF_s) \odot \boldsymbol{\alpha}).$
Here $\text{FFT}(\cdot)$ and $\text{IFFT}(\cdot)$ denote the Fast-Fourier and Inverse-Fast-Fourier transforms, $\odot$ is element-wise multiplication,
and $\balpha(r)\!=\!\tanh(\bs_s)\!+\!1 \; \text{if } r\!<\!r_{\text{thresh}} \; \text{otherwise}\!=\!1$, \noindent where $r$ denotes the radius, and $r_{\text{thresh}}$ refers to the threshold frequency.
Both $\bs_b$ and $\bs_s$ are initially set to zero vectors. 

\par \noindent \textbf{Lightweight in parameters.} The \scaleu module is incorporated into each of UNet's decoder blocks.
It leads to a negligible ($< 0.01\%$) overall increase in the number of parameters and brings noticeable performance gains.

\subsection{\mis}
\label{sec:method-mis}
\figMIS{!t}

To further minimize the information leakage across multiple instance conditionings, we optionally use \mis strategy during the model inference which improves the quality and fidelity of the generated image. 

Specifically, \mis (\cf~\cref{fig:mis}) involves:
1) For each of the $n$ instances, run a separate denoising operation for $M$ steps (less than 10\% of the overall steps) to get the instance latents $L_I$.
Note that, since our model is trained to generate an object within the location token specified for that object, we don't need to explicitly require the model to update the latent representation within the location.
2) Integrate the denoised instance latents $\{L_I^1,\cdots,L_I^n\}$ obtained from step (1) for each of the $n$ objects with the global latent $L_G$, which is derived from all instance tokens and text prompts, by averaging these latents together.
3) Proceed to denoise the aggregated latent from step (2), utilizing all instance tokens and text prompts.

\subsection{Data with Instance Captions}
\label{sec:method-data}

Obtaining a large-scale dataset that contains instance conditions is challenging.
Standard object detection datasets~\cite{lin2014microsoft} only contain a sparse category label, rather than a detailed caption, per object location.
To capture more detailed information about instances and even instance parts, \eg, attributes, we construct a dataset by using multiple models:
\textit{\textbf{1) Image-level label generation}}: We employ RAM~\cite{zhang2023recognize}, a robust open-vocabulary image tagging model, to generate a list of common image-level tags.
\textit{\textbf{2) Bounding-box and mask generation}}: We then use Grounded-SAM~\cite{kirillov2023segany,liu2023grounding} to produce bounding boxes and masks corresponding to these tags.
These tags can at the instance-level, \eg, a parrot, or at the part-level, \eg, a bird's beak. 
\textit{\textbf{3) Instance-level text prompt generation}}: To generate instance-level text prompts that include descriptions of the instances, we crop the instances using their corresponding bounding boxes and create captions for these cropped instances using a pretrained Vision-Language Model (VLM) BLIP-V2~\cite{li2023blip2}.

\subsection{Implementation Details}
\label{sec:impl_details}
We describe salient implementation details and provide the full details in the supplement.

\noindent \textbf{Model training.}
We follow the same settings as GLIGEN~\cite{li2023gligen} and initialize our model with a pretrained \textToI model whose layers are frozen.
We train the model with a batch size of 512 for 100K steps using the Adam optimizer~\cite{kingma2014adam} with a learning rate that is warmed up to $0.0001$ after $5000$ steps.
More details are in appendix materials. 

\noindent \textbf{Training data.}
We automatically generate instance-level masks, boxes and captions following~\cref{sec:method-data}.
We obtain \scribble by randomly sampling points within the masks.
For single-points, we randomly select a point within a circular region of radius $0.1 \cdot r$, centered at the bounding box's center, where $r$ is the length of the shortest side of the box.

\def\tabMainCompare#1{
    \begin{table*}[#1]
    \tablestyle{1.7pt}{1.0}
    \small
    \begin{center}
    \begin{tabular}{lcccclccccclcclcc}
    Location format input $\rightarrow$ & \multicolumn{4}{c}{Boxes} && \multicolumn{5}{c}{Instance Masks} && \multicolumn{2}{c}{Points} && \multicolumn{2}{c}{\Scribble} \\
    \cline{2-5} \cline{7-11} \cline{13-14} \cline{16-17}
    Method & \multicolumn{1}{c}{AP$^{\text{box}}$ } & \multicolumn{1}{c}{AP$_{50}^{\text{box}}$ } & \multicolumn{1}{c}{AR$^{\text{box}}$ } & \multicolumn{1}{c}{FID ($\downarrow$)} &&
    \multicolumn{1}{c}{{IoU} } & \multicolumn{1}{c}{AP$^{\text{mask}}$ } & \multicolumn{1}{c}{AP$_{50}^{\text{mask}}$ } & \multicolumn{1}{c}{AR$^{\text{mask}}$ } & \multicolumn{1}{c}{FID ($\downarrow$)} &&
    \multicolumn{1}{c}{PiM } & \multicolumn{1}{c}{FID ($\downarrow$)} &&
    \multicolumn{1}{c}{PiM} & \multicolumn{1}{c}{FID ($\downarrow$)}
    \\ [.1em]
    \Xhline{0.8pt}
    \textcolor{OracleTextColor}{Upper bound (real images)} & \textcolor{OracleTextColor}{50.2} & \textcolor{OracleTextColor}{66.7} & \textcolor{OracleTextColor}{61.0} & \textcolor{OracleTextColor}{-} && \textcolor{OracleTextColor}{-} & \textcolor{OracleTextColor}{40.8} & \textcolor{OracleTextColor}{63.5} & \textcolor{OracleTextColor}{58.0} & \textcolor{OracleTextColor}{-}&& \textcolor{OracleTextColor}{-}& \textcolor{OracleTextColor}{-}&& \textcolor{OracleTextColor}{-}& \textcolor{OracleTextColor}{-} \\
    \hline
    GLIGEN~\cite{li2023gligen} & 19.6 & 35.0 & 30.7 & 27.0 &&-&-&-&-&-&&-&-&&-&- \\
    GLIGEN~\cite{li2023gligen}$^*$ & 19.3 & 34.6 & 31.1 & - &&-&-&-&-&-&&-&-&&30.2$^{\dagger}$&32.4$^{\dagger}$ \\
    ControlNet~\cite{zhang2023adding}$^\ddagger$ & - & - & - & - && - & 6.5 & 13.8 & 12.9 & - &&-&-&&-&- \\
    DenseDiffusion~\cite{densediffusion} &-&-&-&-&& 35.0 / 48.6 &-&-&-&-&&-&-&&-&- \\
    SpaText~\cite{avrahami2023spatext}$^\ddagger$&-&-&-&-&& - &5.3&12.1&10.7&-&&-&-&&-&- \\
    \hline
    \textbf{\ours} & \bf 38.8 &\bf 55.4 &\bf 52.9 & \bf 23.9 && \bf 61.6 / 71.4 & \bf 27.1 & \bf 50.0 & \bf 38.1 & \bf 25.5 &&\bf 81.1 &\bf 27.5 &&\bf 72.4 &\bf 27.3 \\
    \textit{vs. prev. SoTA} &\plus{+19.2}&\plus{+20.4}&\plus{+21.8}&\plus{-3.1}&&\plus{+25.4} / \plus{+22.8}&\plus{+20.6}&\plus{+36.2}&\plus{+25.2}&-&&-&-&&\plus{+42.2}&\plus{-4.9} \\
    \hline 
    \textcolor{OracleTextColor}{\textbf{\ours} (hybrid)} & \textcolor{OracleTextColor}{\bf 44.6} & \textcolor{OracleTextColor}{\bf 59.6} & \textcolor{OracleTextColor}{\bf 58.8} &\textcolor{OracleTextColor}{\bf 25.5} && - & - & - & - & - && \textcolor{OracleTextColor}{\bf 86.0} &\textcolor{OracleTextColor}{\bf 25.5} &&\textcolor{OracleTextColor}{\bf 82.9} &\textcolor{OracleTextColor}{\bf 26.4} \\
    \end{tabular}
    \end{center}\vspace{-18pt}
    \caption{
    \textbf{Evaluating different location formats} as input when generating images.
    We measure the YOLO recognition performance (AP, AR) for the generated image wrt the location condition provided as inputs, and FID on the \coco \texttt{val} set. 
    Most prior methods only support a handful of the location conditions.
    We observe that \ours, while using the same model parameters, supports various location inputs.
    In each setting, \ours substantially outperforms prior work on all metrics.
    *: evaluated with YOLOv8.
    $^{\dagger}$: GLIGEN's \scribble-based results are derived by using the top-right and bottom-left corners as the bounding box for the region encompassed by the \scribble.
    We measure the IoU using \cite{densediffusion}'s official evaluation codes (left), and YOLOv8-Seg (right).
    $^\ddagger$: ControlNet~\cite{zhang2023adding} (and SpaText~\cite{avrahami2023spatext}) only supports \textit{semantic} segmentation mask inputs, and do not differentiate between instances of the same class. We assess ControlNet's AP$^{\text{mask}}$ using its official mask conditioned Image2Image generation pipeline. Hybrid: we add instance masks as additional conditions.}
    \label{tab:main-results}
    \end{table*}
}

\def\tabAttribute#1{
    \begin{table}[#1]
    \tablestyle{1.3pt}{1.0}
    \small
    \vspace{-8pt}
    \begin{center}
    \begin{tabular}{lcclcclclc}
    \multirow{2}{*}{Methods} & \multicolumn{2}{c}{Color} && \multicolumn{2}{c}{Texture} &&  && \multirow{2}{1.8cm}{\centering Human Eval} \\
    \cline{2-3} \cline{5-6} 
    & Acc$^{\text{color}}$ & CLIP$^{\text{local}}$ && Acc$^{\text{texture}}$ & CLIP$^{\text{local}}$ &&  \\ [.1em]
    \Xhline{0.8pt}
    GLIGEN & 19.2 & 0.206 && 16.6 & 0.206 &&  && 19.7 \\
    \hline
    InstDiff & \bf 54.4 & \bf 0.250 &&\bf 26.8 &\bf 0.225 &&  && 80.3 \\
    $\Delta$ & \plus{+35.2} & \plus{+0.044} && \plus{+10.2} & \plus{+0.019} &&  \\
    \end{tabular}
    \end{center}\vspace{-18pt}
    \caption{\textbf{Attribute binding.}
    We measure whether the attributes of the generated instances match the attributes specified in the instance captions.
    We observe that \ours outperforms prior work on both types of attributes.
    Human evaluators prefer our generations significantly more than the prior work.
    }
    \label{tab:attributes}
    \end{table}
}

\def\tabContribution#1{
    \begin{table}[#1]
    \begin{center}
    \vspace{-8pt}
    \tablestyle{4pt}{1.0}
    \small
    \begin{tabular}{cccccc|lll}
    \rownumber{\#} & {\rotatebox{80}{FA Fusion}} & {\rotatebox{80}{MaskAttn}} & {\rotatebox{80}{ScaleU}} & {\rotatebox{80}{Inst. Cap.}} & {\rotatebox{80}{MIS}} & {{AP$_{50}^{\text{mask}}$ }} & {{Acc$^{\text{color}}$ }} & {FID ($\downarrow$)} \\
    \shline
    \rowcolor{HighlightColor}
    \rownumber{1} & \cmark & \cmark & \cmark & \cmark & \cmark & 50.0 & 55.4 & 25.5 \\
    \rownumber{2} & \gxmark & \cmark & \cmark & \cmark & \cmark & 45.5\diff{5.5} & 49.4\diff{6.0} & 25.8\diff{0.3} \\
    \rownumber{3} &\cmark & \gxmark & \cmark & \cmark & \cmark & 49.3\diff{0.7} & 53.1\diff{2.3} & 25.7\diff{0.2} \\
    \rownumber{4} & \cmark & \cmark & \gxmark & \cmark & \cmark & 47.7\diff{2.3} & 52.2\diff{3.2} & 25.7\diff{0.2} \\
    \rownumber{5} &\cmark & \cmark & \cmark & \gxmark & \cmark & 47.8\diff{2.2} & 38.2\diff{17.2} & 25.6\diff{0.1} \\
    \rownumber{6} &\cmark& \cmark & \cmark & \cmark & \gxmark & 49.8\diff{0.2} & 49.5\diff{5.9} & 28.6\diff{3.1} \\
    \end{tabular}
    \end{center}\vspace{-18pt}
    \caption{\textbf{Contribution of each component} evaluated by removing or adding it and measuring the impact of the generated image in terms of its instance location performance (AP), and instance attribute binding (Acc), and overall image quality (FID). When Format Aware (FA) fusion mechanism is disabled, we use the Joint format fusion mechanism instead.
    \bgcolortext{Top row} is the default setting for \ours in the paper and we report the drop in performance for each subsequent row in \textcolor{DifferenceColor}{\textbf{red}}.
    }
    \label{tab:abl-components}
    \end{table}
}

\def\tabHybridInputs#1{
    \begin{table}[#1]
    \begin{center}
    \vspace{-3pt}
    \tablestyle{4.4pt}{1.0}
    \small
    \begin{tabular}{ccc|cccccc}
    {point} & box & mask & PiM & {AP$^{\text{box}}$} & {AP$_{50}^{\text{box}}$ } & {AP$^{\text{mask}}$} & {AP$_{50}^{\text{mask}}$ } \\
    \shline
    \cmark & \gxmark & \gxmark & 81.1 & - & - & - & - \\
    \cmark & \cmark  & \gxmark & 85.6 & 38.8 & 55.4 & - & - \\
    \cmark & \cmark  & \cmark  &\bf 86.0 &\bf 44.6 &\bf 59.6 & \bf 27.1 & \bf 50.0 \\
    \end{tabular}
    \end{center}\vspace{-16pt}
    \caption{\textbf{Multiple location formats at inference} improves performance and helps the model to better respect location conditions.
    }
    \label{tab:abl-hybrid-points}
    \end{table}
}

\def\tabMIS#1{
    \begin{table}[#1]
    \vspace{-4pt}
    \begin{center}
    \tablestyle{3.5pt}{1.0}
    \small
    \begin{tabular}{c|cc|cccccc}
    & GLIGEN~\cite{li2023gligen} & w/ MIS & InstanceDiffusion & w/ MIS \\
    \shline
    Acc$^{\text{color}}$ & 19.2 &\bf 29.7 & 49.5 &\bf 55.4 \\
    \end{tabular}
    \end{center}\vspace{-18pt}
    \caption{\textbf{\mis} can be adapted for previous location conditioned work, yielding notable performance gains. 
    }
    \label{tab:mis}
    \end{table}
}

\def\tabMISSteps#1{
    \begin{table}[#1]
    \vspace{-4pt}
    \begin{center}
    \tablestyle{4pt}{1.0}
    \small
    \begin{tabular}{cccccccc}
    \% of Steps $\rightarrow$ & 0\% & 10\% & 20\% & 30\% & 36\% & 40\% & 50\% \\ [.1em]
    \shline
    FID  & 28.6 & 27.8 & 27.4 & 25.8 & 25.5 & 25.0 & 27.0 \\
    AP$^{\text{mask}}_{50}$ & 49.8 & 49.8 & 49.4 & 49.4 & 50.0 & 49.2 & 48.3 \\
    \end{tabular}
    \end{center}\vspace{-18pt}
    \caption{\textbf{\mis} (MIS) lowers the FID and improves overall image quality. Location conditions: instance masks. 
    }
    \label{tab:missteps}
    \end{table}
}

\def\tabAblations#1{
\begin{table*}[#1]
	\centering
        \vspace{-6pt}
        \subfloat[
	\textbf{ScaleU}
	\label{tab:ablate_scaleu}
	]{
            \begin{minipage}{0.22\linewidth}{\begin{center}
                \tablestyle{1.1pt}{1.2}
                \begin{tabular}{ccaH}
                    versions $\rightarrow$ & FreeU~\cite{si2023freeu} & ScaleU & SE-ScaleU \\ [.1em]
                    \shline
                    AP$^{\text{box}}_{50}$ & 52.2 & 55.4 & 55.2 \\
                \end{tabular}
            \end{center}}\end{minipage}
	}
        \subfloat[
            \textbf{extra tokens from binary masks}
            \label{tab:ablate_tokens}
            ]{
                \centering
                \begin{minipage}{0.32\linewidth}{
                \begin{center}
                \tablestyle{1.1pt}{1.2}
                \begin{tabular}{cca}
                    methods $\rightarrow$ & w/o extra tokens & w/ extra tokens \\ [.1em]
                    \shline
                    AP$^{\text{mask}}_{50}$ & 46.7 & 50.0 \\
                \end{tabular}
            \end{center}}\end{minipage}
        }
        \subfloat[
        \textbf{mask parameterization}
        \label{tab:ablate_mask_type}
        ]{
            \centering
            \begin{minipage}{0.2\linewidth}{
            \begin{center}
            \tablestyle{1.2pt}{1.2}
            \begin{tabular}{cca}
                format $\rightarrow$ & polygons & +inside \\ [.1em]
                \shline
                AP$^{\text{mask}}_{50}$ & 47.5 & 50.0 \\
            \end{tabular}
            \end{center}}\end{minipage}
        }
        \subfloat[
	\textbf{\texttt{\#} points per mask}
	\label{tab:ablate_n_point_mask}
	]{
		\centering
		\begin{minipage}{0.22\linewidth}{
            \begin{center}
                \tablestyle{1.2pt}{1.2}
                \begin{tabular}{cccac}
                    \# points $\rightarrow$ & 64 & 128 & 256 & 512 \\ [.1em]
                    \shline
                  AP$^{\text{mask}}_{50}$ & 45.7 & 48.5 & 50.0 & 50.0 \\
                \end{tabular}
		\end{center}}\end{minipage}
	}
    \vspace{-8pt}
    \caption{\textbf{Ablating design choices} where the default settings are indicated in \colorrowtext{}.
    \textbf{(a)} Compared to FreeU, our proposed \scaleu block improves the models ability to respect location conditions.
    \textbf{(b)} Using extra tokens from binary instances masks can improve the mask AP.
    \textbf{(c)} Parameterizing the instance masks using points on their boundaries and inside is beneficial.
    \textbf{(d)} Increasing the number of points used to parameterize masks improves performance.
    }
    \label{tab:ablate}
\end{table*}
}

\def\tabLVIS#1{
    \begin{table}[#1]
    \tablestyle{1.6pt}{1.0}
    \small
    \vspace{-6pt}
    \begin{center}
    \begin{tabular}{lcccccccc}
    Methods & AP & AP$_{50}$ & AP$_s$ & AP$_m$ & AP$_l$ & AP$_r$ & AP$_c$ & AP$_f$ \\ [.1em]
    \Xhline{0.8pt}
    \textcolor{OracleTextColor}{Upper bound} & \textcolor{OracleTextColor}{44.6} & \textcolor{OracleTextColor}{57.7} & \textcolor{OracleTextColor}{33.2} & \textcolor{OracleTextColor}{55.0} & \textcolor{OracleTextColor}{66.1} & \textcolor{OracleTextColor}{31.4} & \textcolor{OracleTextColor}{44.5} & \textcolor{OracleTextColor}{50.5}\\
    \hline
    GLIGEN~\cite{li2023gligen}$^{\dagger}$ & 9.9 & 9.5 & 1.6 & 10.5 & 31.1 & 7.4 & 10.0 & 10.9 \\
    \hline
    InstanceDiffusion & 17.9 & 25.5 & 5.5 & 24.2 & 45.0 & 12.7 & 18.7 & 19.3 \\
    \textit{vs. prev. SoTA} &\plus{+8.0}&\plus{+16.0}&\plus{+3.9}&\plus{13.7}&\plus{+13.9}&\plus{+5.3}&\plus{+8.7}&\plus{+8.4} \\
    \end{tabular}
    \end{center}\vspace{-18pt}
    \caption{\textbf{Box inputs on LVIS} \texttt{val}. We evaluate using a pretrained detector (ViTDet-L~\cite{li2022exploring}) and obtain the upper bound by evaluating the detector on real images resized to 512$\times$512.
    \ours significantly outperforms prior work across all metrics including object sizes, and class frequencies. $^{\dagger}$: reproduced results.
    }
    \label{tab:lvis}
    \end{table}
}

\def\figDemoGLIGEN#1{
    \captionsetup[sub]{font=small}
    \begin{figure*}[#1]
      \centering
      \vspace{-3.5pt}
      \includegraphics[width=0.98\linewidth]{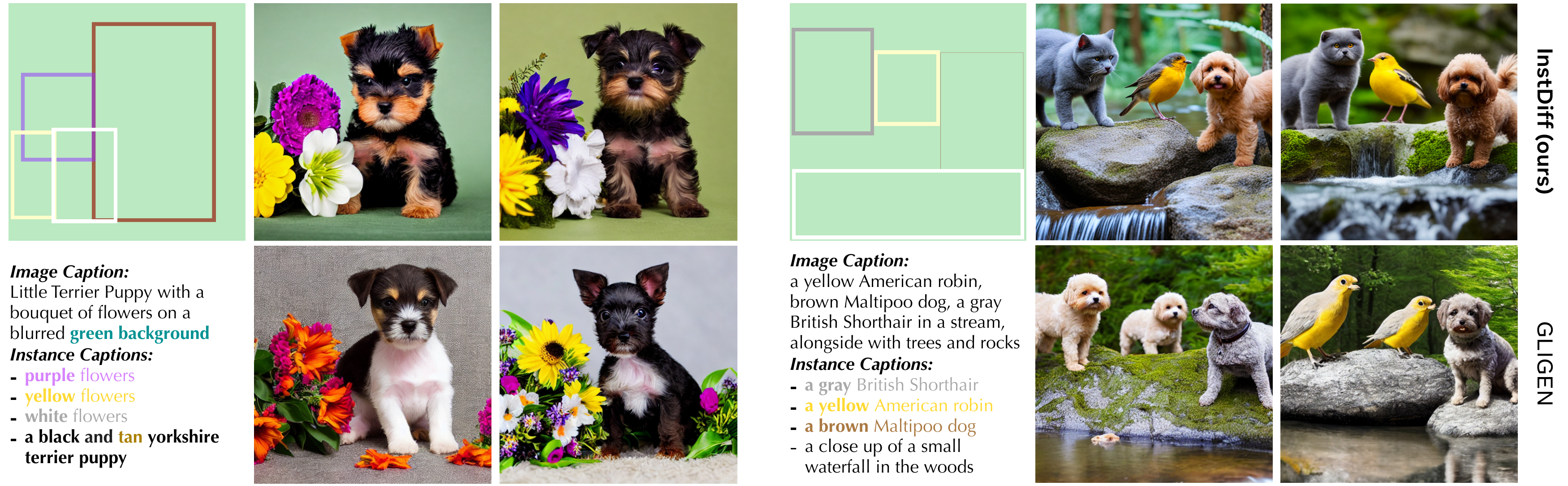}\vspace{-6pt}
      \caption{\textbf{Qualitative comparison of \ours \vs GLIGEN} conditioned on multiple instance boxes and prompts.
      Prior work (bottom row) fails to accurately reflect specific instance attributes, \eg, colors for the flower and puppies on the left, and not depicting a waterfall on the right.
      The generations also do not capture the correct instances, and are prone to information leakage across the instance prompts, \eg, generating two similar instances on the right. \ours effectively mitigates these issues.
      }
      \label{fig:demo-vs-gligen}
    \end{figure*}
}

\def\figHybridInputs#1{
    \captionsetup[sub]{font=small}
    \begin{figure}[#1]
      \vspace{-8pt}
      \centering
      \includegraphics[width=0.98\linewidth]{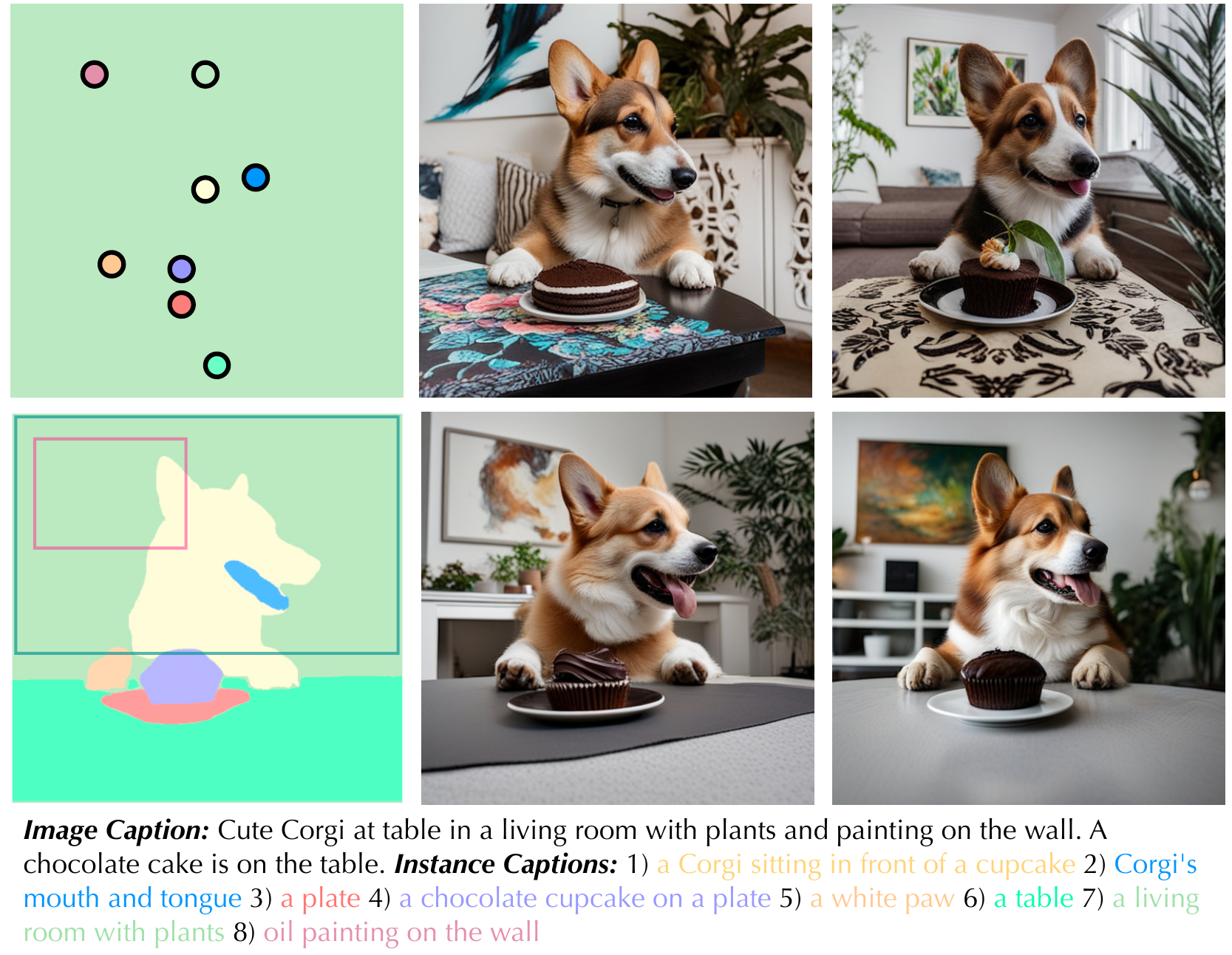}\vspace{-8pt}
      \caption{\ours image generation using various location conditions: points (row 1) and masks (row 2).}
      \label{fig:demo3}
    \end{figure}
}

\def\figIterative#1{
    \captionsetup[sub]{font=small}
    \begin{figure}[#1]
      \centering
      \includegraphics[width=0.98\linewidth]{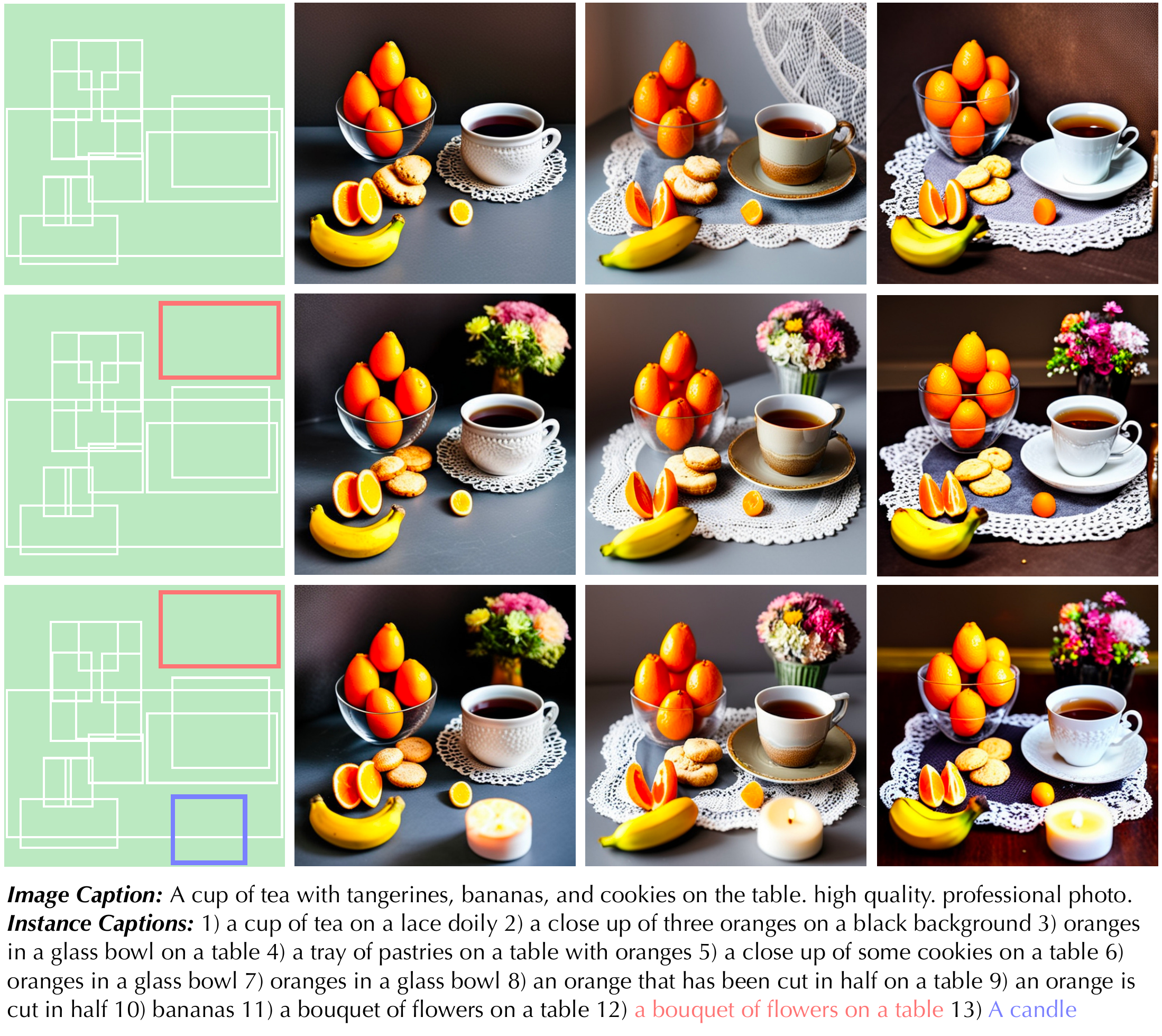}\vspace{-8pt}
      \caption{\ours can also support \textbf{iterative image generation}. 
      Using the identical initial noise and image caption, \ours can progressively add new instances (like a bouquet of flowers in row two and a candle in row three), while minimally altering the pre-generated instances (row one). More results on iterative image generation that supports instance editing, replacing, moving and resizing can be found in appendix materials. 
      }
      \label{fig:demos-edit}
    \end{figure}
}


\section{Experiments}
\label{sec:experiments}

\subsection{Experimental setup}
\label{sec:exp-setup}

\textbf{Training data.}
Prior work, notably GLIGEN~\cite{li2023gligen}, relies on automatic annotations that use open-vocabulary detection models. These do not yield per-instance captions and different location formats such as \scribble \etc (Note: `mask' conditioning in prior work~\cite{li2023gligen,avrahami2023spatext} is per-category and not per-instance).
Thus, to support the richer conditioning proposed in our work, we rely on recognition models as described in~\cref{sec:method-data,sec:impl_details} to generate instance-level annotations include different location formats (masks, boxes, \scribble{}s, single-points) and per-instance captions.
To ensure fair comparison to prior work~\cite{li2023gligen}, we use approximately the same number of images (5M) from an internal licensed dataset of natural images and paired global text.

\noindent \textbf{Test data.}
We use standard benchmarks with bounding box and instance masks: 1) COCO~\cite{lin2014microsoft} \texttt{val} with 80 classes; 2) large vocabulary instance segmentation dataset LVIS~\cite{gupta2019lvis} \texttt{val} with over 1200 classes; 3) 250 selected samples ($\sim$2 objects per image) from COCO \texttt{val} as in~\cite{densediffusion}.
We do not use the real images from the dataset, and only use the text and location conditions.
Notably, we also do not use any information from the \texttt{train} splits of the data which makes our evaluations zero-shot.

\label{sec:exp-metrics}

\tabMainCompare{t}
\figDemoGLIGEN{t}

\par \noindent \textbf{Evaluation metrics for alignment to instance locations.}
We measure how well the objects in the generated image adhere to different location formats in the input.
\par \noindent \textit{\textbf{Bounding box.}}
We follow prior work~\cite{redmon2016you,Jocher_YOLO_by_Ultralytics_2023,li2023gligen,densediffusion} and use the YOLO score.
Specifically, we use a pretrained YOLOv8m-Det~\cite{Jocher_YOLO_by_Ultralytics_2023} detection model.
We compare the model's detected bounding boxes on the generated image with the bounding boxes specified in the input using COCO's official evaluation metrics (AP and AR).
We report AP$^{\text{box}}_\text{l}$, AP$^{\text{box}}_\text{m}$, and AP$^{\text{box}}_\text{s}$, which evaluate the model's performance based on different object sizes.
\par \noindent \textit{\textbf{Instance mask.}}
We compare YOLOv8m-Seg~\cite{Jocher_YOLO_by_Ultralytics_2023}'s detected instance masks in the generated image to the masks specified in the input using the COCO AP and AR metrics.
To compare with~\cite{densediffusion}, we report the IOU score for the mask.
\par \noindent \textit{\textbf{\Scribble.}}
Since prior work has not reported on alignment performance using \scribble, we introduced a new evaluation metric using YOLOv8m-Seg.
We report {``Points in Mask''} (\textbf{PiM}), which measures how many of randomly sampled points in the input \scribble lie within the detected mask.
\par \noindent \textit{\textbf{Single-point.}}
Similar to \scribble, the instance-level accuracy \textbf{PiM} is 1 if the input point is within the detected mask, and 0 otherwise.
We then calculate the averaged \textbf{PiM} score.

\par \noindent \textbf{\textbf{Evaluation metrics for alignment to instance prompts.}}
We measure the alignment of the objects in the generated image to the corresponding text and location conditions from \coco \texttt{val} set.

\par \noindent \textit{\textbf{Compositional attribute binding.}}
We measure if the generated instances adhere to the attribute (color and texture) specified in the instance prompts.
We use YOLOv8-Det to detect the bounding boxes.
We feed the cropped box to the CLIP model to predict its attribute (colors and textures), and measure the accuracy of the prediction with respect to the attribute specified in the instance prompt.
We use 8 common colors, \ie, ``black'', ``white'', ``red'', ``green'', ``yellow'', ``blue", ``pink'', ``purple'', and 8 common textures, \ie, ``rubber'', ``fluffy'', ``metallic'', ``wooden'', ``plastic'', ``fabric'', ``leather'' and ``glass''.

\noindent\textit{\textbf{Instance text-to-image alignment}}:
We report the CLIP-Score on cropped object images (Local CLIP-score~\cite{radford2021learning,avrahami2023spatext}), which measures the distance between the instance text prompt's features and the cropped object images.

\noindent\textit{\textbf{{Global text-to-image alignment:}}}
CLIP-Score~\cite{radford2021learning,rombach2022high} between the input text prompt and the generated image.

\noindent{\textbf{Human evaluation}:
We evaluate both the fidelity wrt instance-level conditions (locations and text prompts) and the overall aesthetic of the generated images.
We prompt users to select results that more closely adhere to the provided layout conditions and the accompanying instance captions. 
This evaluation is conducted on 250 samples, each accompanied by instance-level captions and bounding boxes.

\subsection{Comparison with prior work}
\label{sec:exp-results}

\tabAttribute{!t}
\tabLVIS{!t}

\noindent \textbf{Single location format at inference.}
We assess the efficacy of multiple methods in generating images under diverse location formats and report results in~\cref{tab:main-results}.
Since our evaluation uses recognition model (YOLO), we establish an upper bound by measuring the recognition performance on the real dataset images corresponding to the text and location conditions.
Overall, our results show that \ours outperforms all prior work across various location conditions when measured across all evaluation metrics for object location and image quality.
Next, we discuss the results for each location format.
\underline{Box input}:
\ours achieves the highest AP$^\text{box}$ of 38.8 and AR$^\text{box}$ of 52.9, outperforming the previous state-of-the-art by a significant margin, +19.2 and +21.8 for AP$^\text{box}$ and AR$^\text{box}$, respectively.
The reduction in FID score for \ours demonstrates its ability to produce high-quality images while adhering to the prescribed location conditions.
\underline{Instance mask input} imposes stricter constraints on the instance location than box input and is more challenging than the semantic masks studied in prior work~\cite{li2023gligen,zhang2023adding} that do not distinguish individual instances.
Even in this challenging setting, \ours outperforms prior SOTA \cite{densediffusion} significantly.
\underline{Points and \Scribble}:
Given the lack of prior studies that present quantifiable results for these location inputs, we introduce these novel evaluation metrics and benchmarks, establishing a new baseline for future research endeavors.
Note that the term `scribble' in ControlNet~\cite{zhang2023adding} refers to object boundary sketches rather than scribbles used in our work which follows ~\cite{lin2016scribblesup,bai2014error,agustsson2019interactive}.

\noindent \textbf{Attribute binding}.
In~\cref{tab:attributes}, we measure whether the attributes (color and texture) of the generated instances match the attributes specified in the instance captions.
We observe that attribute binding is challenging for the prior SOTA method, GLIGEN while \ours significantly improves on both color and texture binding.
Adhering to texture seems to be more challenging than colors, \eg, \texttt{wooden dog} \vs. \texttt{red dog}, as reflected by the lower accuracies for all methods on this task.
We compare the generations produced by both models using human evaluators and find that humans strongly prefer our generations over prior work (80.3\% preference) confirming their high generation quality and controllability.

\noindent \textbf{Challenging box inputs}.
In~\cref{tab:lvis}, we evaluate zero-shot performance on the challenging LVIS~\cite{gupta2019lvis} dataset which has $15\times$ more classes than \coco, and many more instances per sample ($\sim$12 objects per images).
Even on this challenging dataset, \ours outperforms prior work across all metrics.
The gain is particularly strong on medium to large sized objects.

\figHybridInputs{!t}

\noindent \textbf{Multiple location formats at inference} are analyzed in~\cref{tab:abl-hybrid-points}.
We observe that using all formats together provides the best performance and more precise control on the instance location.
This confirms the benefit of our design choice to model all location formats.

\tabHybridInputs{!t}
\tabContribution{!t}
\tabAblations{!t}
\noindent \textbf{Qualitative results.} \cref{fig:demo-vs-gligen} provides qualitative comparisons between \ours and the previous SOTA method, GLIGEN~\cite{li2023gligen}, when given multiple instance boxes and associated text prompts.
We see that GLIGEN often misinterprets specific instance attributes; \eg, it incorrectly renders the colors of flowers and puppies on the left, and fails to produce a waterfall in the right images.
GLIGEN also shows `information leakage' across instance prompts (generating duplicate birds for the two images on the right).
In~\cref{fig:demo3}, we show more qualitative results using different location conditions for \ours.

\tabMISSteps{!t}
\tabMIS{!t}
\subsection{Ablation study}
\label{sec:exp-abl}
We ablate the components in \ours and use the \coco \texttt{val} set and provide mask, box and point location formats per-instance as input by default.
\noindent \textbf{Some design choices} used in our method are ablated in in~\cref{tab:ablate}.
We compare our proposed \scaleu block with FreeU in~\cref{tab:ablate_scaleu}.
\scaleu leads to an improved localization AP suggesting that our learnable scaling of the backbone features outperforms the manually tuned FreeU.
The impact of using extra tokens generated from binary instance masks is explored in \cref{tab:ablate_tokens}.
Lastly, for mask-conditioned input, \cref{tab:ablate_n_point_mask,tab:ablate_mask_type} show that points derived from both polygons and instance masks and using 128 points per instance mask gives the optimal performance.

\noindent \textbf{Contribution of each component} its effect on image generation is measured in~\cref{tab:abl-components}.
We compare using different design choices for the fusion mechanism in \unifusion that fuse the location condition embeddings with the backbone \textToI features: Format Aware fusion (row \rownumber{1}) or the Joint Format fusion (row \rownumber{2}).
We find that making the fusion mechanism format-aware significantly improves performance since the location formats specify varying degrees of control on the instance location.
Comparing rows \rownumber{1, 3} shows that using \maskedattn for fusing the location features helps the model focus on instance-specific regions and thus improves attribute binding (color accuracy).
Removing \scaleu (rows \rownumber{1, 4}) causes a significant drop in AP$_{50}^\text{mask}$ and Acc$^{\text{color}}$ scores.
This underscores the importance of dynamically adjusting the channel weights of both skip connected and backbone features.
In row \rownumber{5}, we observe that our generated instance captions are critical for learning attribute binding, as indicated by the 17\% drop in Acc$^{\text{color}}$ after removing them.
Finally, row \rownumber{6} shows that \mis (MIS) improves the overall image quality (lower FID) and attribute binding (color accuracy).

\figIterative{!t}

\par \noindent \textbf{\mis}
The impact of the proportion of MIS steps used in inference is explored inn ~\cref{tab:missteps}. 
MIS can effectively improve the quality of the generated images and attribute binding when the MIS percentage is below 36\%.
As shown in~\cref{tab:mis}, we applied \mis to other location-conditioned \textToI models and observed significant gains for the attribute binding ability of GLIGEN.
These results confirm that MIS minimizes information leakage and that it can be easily used to improve other location-conditioned models.

\par \noindent \textbf{Application: Iterative generation.}
Since \ours allows for precise control over the instances, we show a useful application that benefits from this property in~\cref{fig:demos-edit}.
\ours allows users to selectively insert objects into precise locations while preserving the integrity of previously generated objects and the global scene.
We hope that the precise control enabled by \ours will lead to many other such useful applications.

\section{Conclusions, Limitations and Future Work}
We presented \ours which enables precise instance-level control for \textToI generation and significantly outperforms all prior work in terms of complying with instance attributes and accommodates a variety of location formats -- masks, boxes, \scribble{}s and points.
Our studies indicate that there is a noticeable disparity in the generation quality of small objects compared to larger ones.
We also find that texture binding for instances poses a challenge across all methods tested, including \ours. 
Improving instance conditioning for these cases is an important direction for future research.

{
\small
\bibliographystyle{ieeenat_fullname}
\bibliography{main}
}

\clearpage
\renewcommand{\thefigure}{A\arabic{figure}}
\setcounter{figure}{0}
\renewcommand{\thetable}{A\arabic{table}}
\setcounter{table}{0}
\renewcommand{\thesection}{A\arabic{section}}
\setcounter{section}{0}
\maketitlesupplementary

\def\figDemoSteps#1{
    \captionsetup[sub]{font=small}
    \begin{figure*}[#1]
      \centering
      \includegraphics[width=1.0\linewidth]{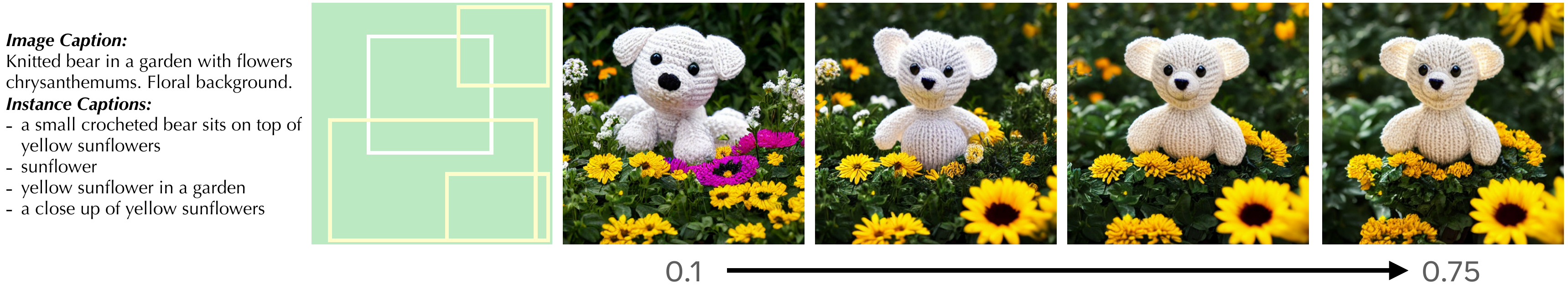}
      \caption{As the \unifusion module is integrated for an increasing proportion of timesteps (from 5\% timesteps to 75\% timesteps), the model's adherence to the instance conditions progressively improves. 
      The generation of the sunflower at the top left corner occurs once the UniFusion module is activated for 75\% of the total timesteps.}
      \label{fig:unifuion_steps}
    \end{figure*}
}

\def\figMoreDemos#1{
    \captionsetup[sub]{font=small}
    \begin{figure*}[#1]
      \centering
      \includegraphics[width=1.0\linewidth]{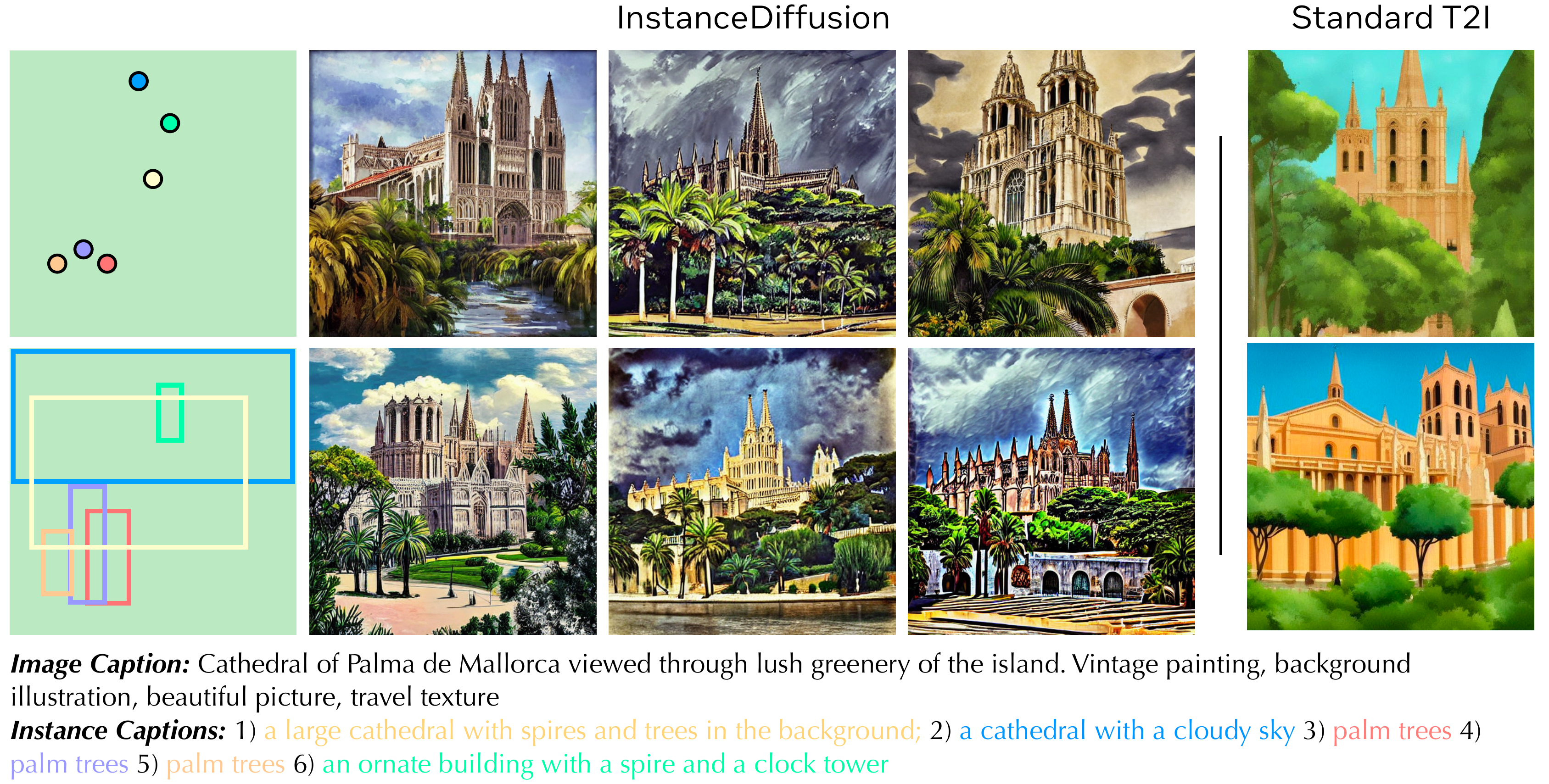}\vspace{20pt}
      \includegraphics[width=1.0\linewidth]{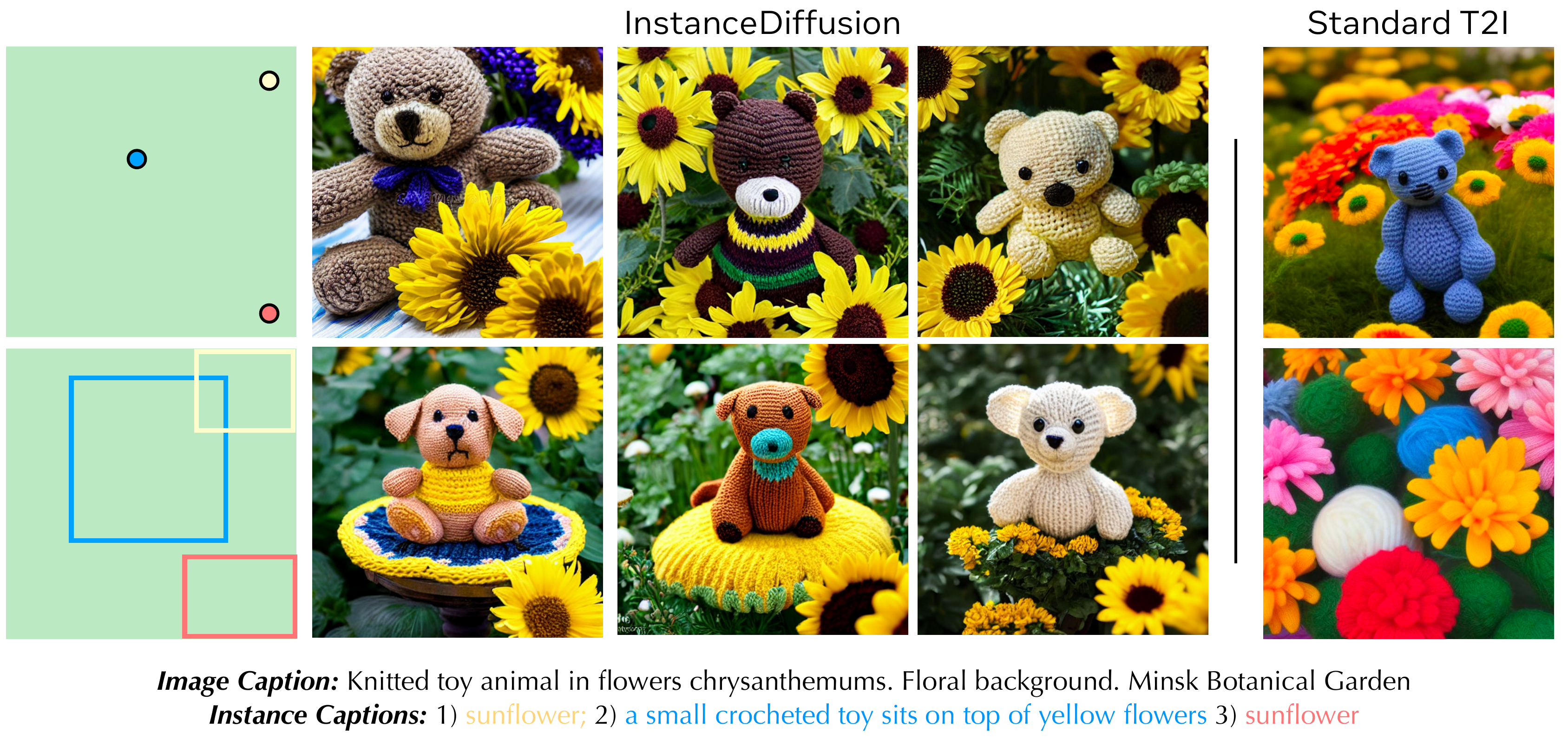}
      \caption{More demo images on image generation with point and bounding box as model inputs. The standard Text-to-Image model refers to the pretrained text-to-image model \ours and GLIGEN used. Standard T2I model uses the image caption as the model input to generates these images.}
      \label{fig:demo_supp_ex2}
    \end{figure*}
}

\def\figDemoPointScribble#1{
    \captionsetup[sub]{font=small}
    \begin{figure*}[#1]
      \centering
      \includegraphics[width=1.0\linewidth]{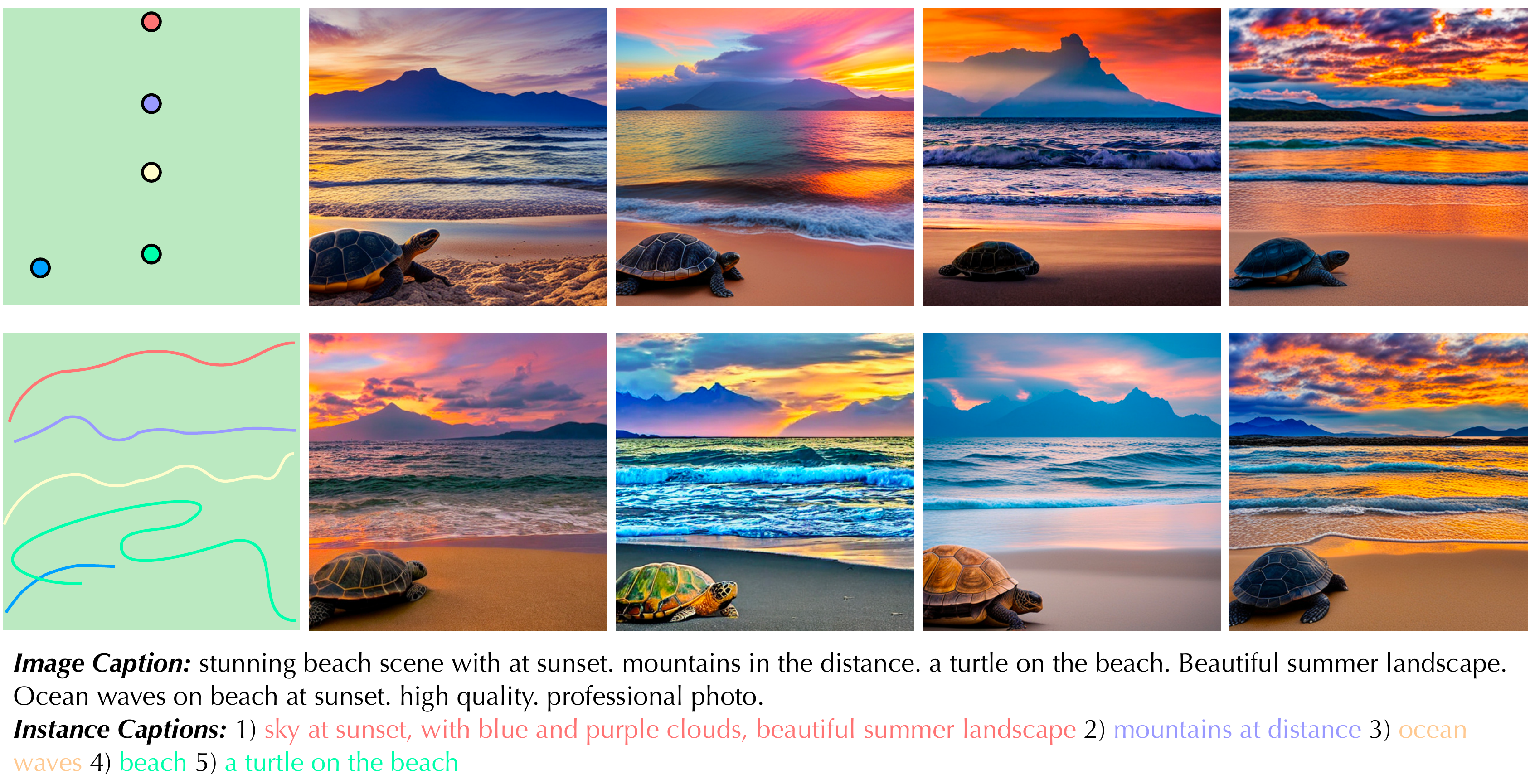}\vspace{20pt}
      \includegraphics[width=1.0\linewidth]{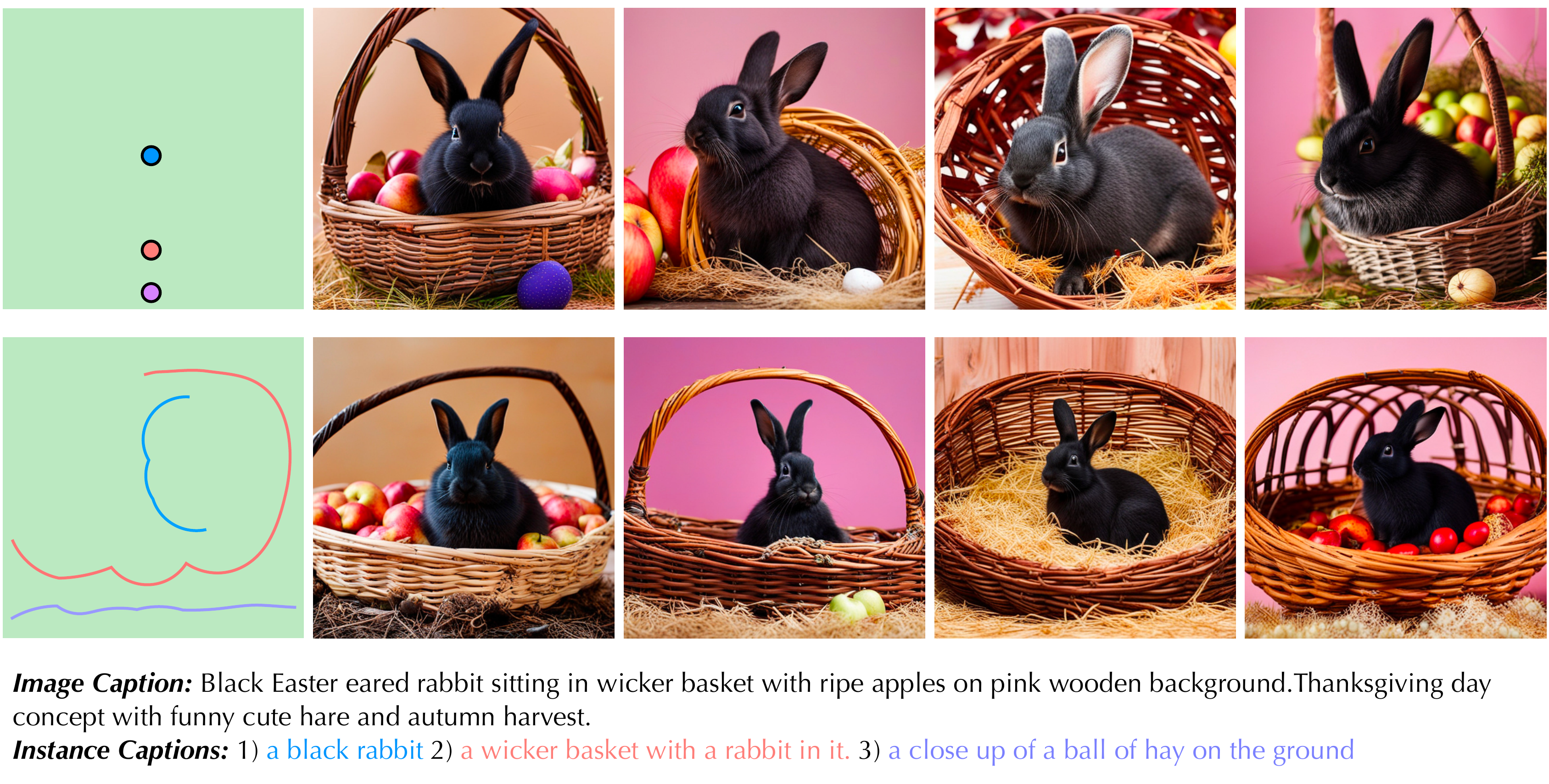}
      \caption{More image generations with point and scribbles as model inputs, which were not supported by previous layout conditioned text-to-image models.}
      \label{fig:demo_supp_ex1}
    \end{figure*}
}

\def\figDemoPose#1{
    \captionsetup[sub]{font=small}
    \begin{figure*}[#1]
      \centering
      \includegraphics[width=1.0\linewidth]{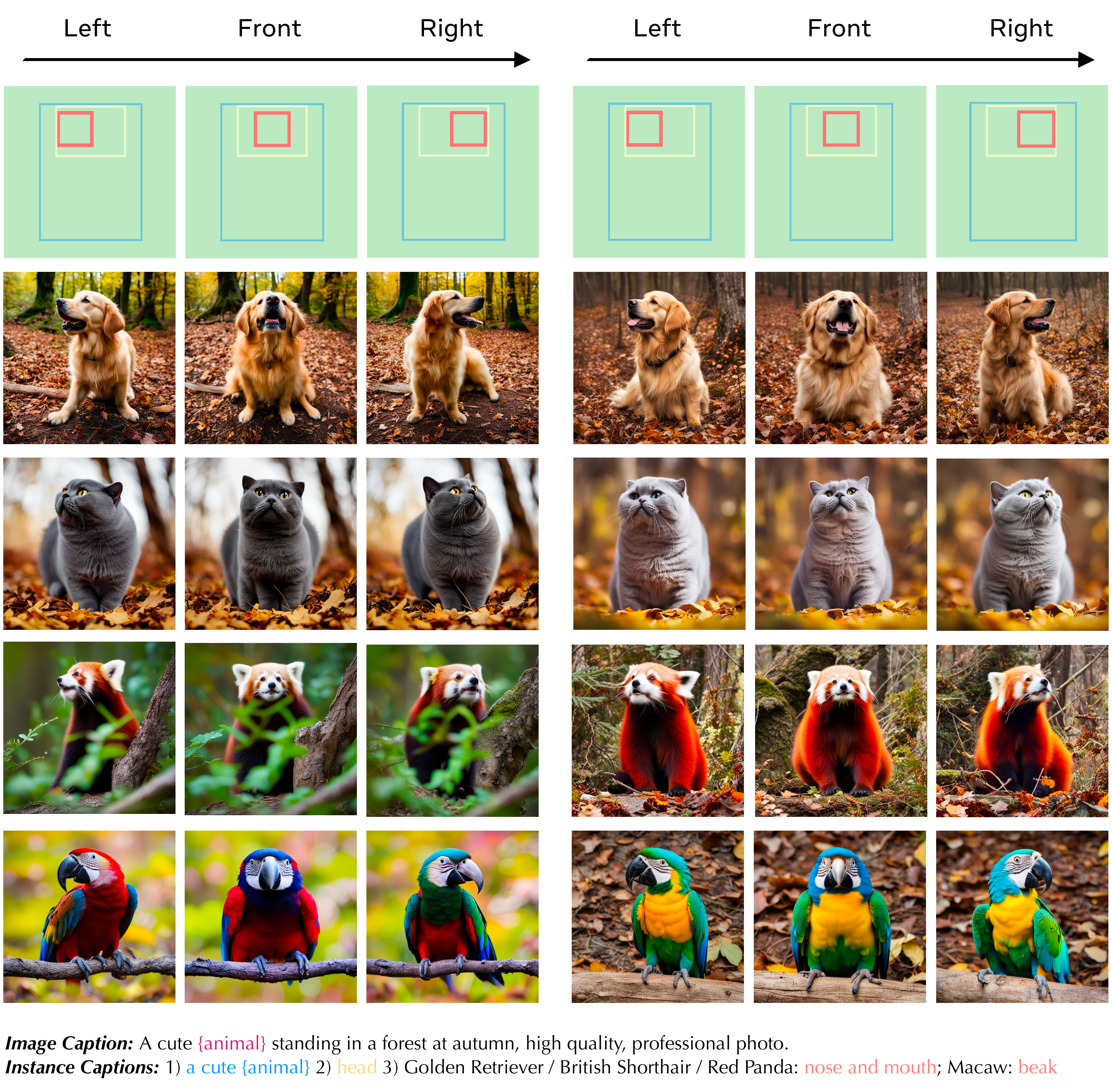}
      \caption{Let's get everybody turning heads! Hierarchical location conditioning in image composition. These results illustrate how the orientation of parts and subparts subtly influences the pose of the whole object (right, left, front), demonstrating the application of spatial hierarchy in visual design. We anticipate that this capability will pave the way for further research and applications in achieving more precise control in image generation. }
      \label{fig:demo_pose}
    \end{figure*}
}

\def\tabMISDesign#1{
\begin{table}[#1]
    \centering
    \tablestyle{5pt}{1.2}
    \begin{tabular}{cccacc}
        & crop-and-paste & latents averaging \\ [.1em]
        \shline
        FID & 24.3 &\bf 23.9 \\
        \hline
        AP$_{50}^{\text{mask}}$ & 49.1 &\bf 50.0 \\
    \end{tabular}
    \caption{Model inference with \mis using different \mis design variations.}
    \label{tab:ablate_mix_design}
\end{table}
}

\def\tabAblationsUniFusion#1{
\begin{table}[#1]
    \centering
    \subfloat[
    \textbf{freq. bandwidth}
    \label{tab:ablate_f_band_sup}
    ]{
        \begin{minipage}{0.45\linewidth}{\begin{center}
                    \tablestyle{1.1pt}{1.2}
                    \begin{tabular}{cccacc}
                        Bandwidth $\rightarrow$ & 4 & 8 & 16 & 32 \\ [.1em]
                        \shline
                        AP$^{\text{box}}_{50}$ & 50.8 & 53.9 & 55.4 & 55.3 \\
                    \end{tabular}
        \end{center}}\end{minipage}
    }
    \subfloat[
    \textbf{MLP dim}
    \label{tab:ablate_unifusion_dim_sup}
    ]{
        \begin{minipage}{0.45\linewidth}{\begin{center}
                    \tablestyle{1.1pt}{1.2}
                    \begin{tabular}{cccac}
                        $N$ $\rightarrow$ & 512 & 2048 & 3072 & 4096 \\ [.1em]
                        \shline
                        AP$^{\text{box}}_{50}$ & 52.9 & 53.5 & 55.4 & 55.4 \\
                    \end{tabular}
        \end{center}}\end{minipage}
    }
    \caption{
        \textbf{Ablating design choices for \unifusion.} Components and default settings are highlighted in \colorrowtext{}.
        \textbf{(a)} We vary the frequency bandwidth used in the Fourier embeddings of the point coordinates in the \unifusion block.
        \textbf{(b)} We study the impact of the dimensionality of MLP layers in the \unifusion block.}
    \label{tab:ablate_appendix}
\end{table}
}

\def\tabAblationsScaleU#1{
\begin{table}[#1]
    \centering
    \tablestyle{5pt}{1.2}
    \begin{tabular}{ccac}
        Versions $\rightarrow$ & FreeU~\cite{si2023freeu} & ScaleU & SE-ScaleU \\ [.1em]
        \shline
        AP$^{\text{box}}_{50}$ & 52.2 & 55.4 & 55.2 \\
    \end{tabular}
    \caption{
    We evaluate the performance of the lightweight \scaleu (\cref{fig:scaleu-all} b) against the dynamically adaptable SE-ScaleU (\cref{fig:scaleu-all} c), and further compare our \scaleu with FreeU~\cite{si2023freeu}, a previous work that manually tune the scaling vectors.}
    \label{tab:ablate_appendix_scaleu}
\end{table}
}

\def\tabHybridInputs#1{
    \begin{table}[#1]
    \begin{center}
    \tablestyle{3pt}{1.2}
    \small
    \begin{tabular}{ccc|cc|ccc|cc}
    box & point & mask & \multicolumn{1}{c}{AP$^{\text{box}}$ } & \multicolumn{1}{c|}{AP$_{50}^{\text{box}}$ } & point & box & mask & \multicolumn{1}{c}{PiM} \\
    \shline
    \cmark & \gxmark & \gxmark & 36.1 & 52.4 &\ccolor \cmark &\ccolor \gxmark &\ccolor \gxmark &\ccolor 79.7 \\
    \ccolor \cmark & \ccolor \cmark  &\ccolor \gxmark &\ccolor 38.8	&\ccolor 55.4 & \cmark & \cmark  & \gxmark & 85.6 \\
    \cmark & \cmark  & \cmark  & \bf 44.6 &\bf 59.6 & \cmark & \cmark  & \cmark &\bf 86.0 \\
    \shline
    mask & box & point & \multicolumn{1}{c}{AP$^{\text{mask}}$ } & \multicolumn{1}{c|}{AP$_{50}^{\text{mask}}$ } & \scribble & box & mask & \multicolumn{1}{c}{PiM} \\
    \shline
    \cmark & \gxmark & \gxmark & 13.6 & 27.3 &\ccolor \cmark &\ccolor \gxmark &\ccolor \gxmark &\ccolor 72.4 \\
    \cmark & \cmark  & \gxmark  & 20.9 & 40.9 & \cmark & \cmark  & \gxmark & 74.8 \\
    \ccolor \cmark &\ccolor \cmark  &\ccolor \cmark   &\ccolor \bf 24.6 &\ccolor \bf 50.0 & \cmark & \cmark  & \cmark &\bf 82.9 \\
    \end{tabular}
    \end{center}\vspace{-4pt}
    \caption{\textbf{Model inference with hybrid location inputs.} We found that hybrid inputs can often help the model to better respect the location conditions and lead to performance gains.
    Default inference setting is colored in \textcolor{Gray}{gray}.
    \textit{{Note: Given a box, one can always determine a point by using its center. Similarly, from a mask, both a box and a central point can be derived without the need for extra user inputs.}
    }
    }
    \label{tab:abl-hybrid}
    \end{table}
}

\def\figScaleUDiagram#1{
    \captionsetup[sub]{font=small}
    \begin{figure*}[#1]
    \centering
    \includegraphics[width=1.0\linewidth]{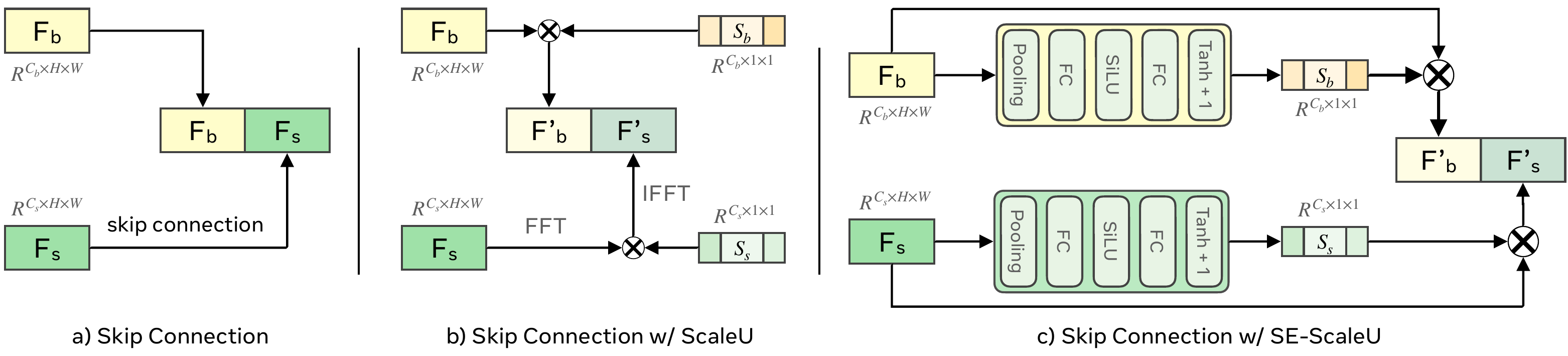}
    \caption{
    Various design choices for the {\scaleu block}. In the UNet architecture, $\bF_b$ represents the main features, while $\bF_s$ denotes the skip connected features. 
    Typically, UNet employs skip connections as shown in (a) to pass features from the encoder to the decoder, aiding in recovering spatial information lost in downsampling. 
    We introduce ScaleU (b), which re-calibrates both the main and skip-connected features prior to their concatenation. 
    Additionally, we implement SE-ScaleU (c), which utilizes an MLP layer—akin to the Squeeze-and-Excitation module~\cite{hu2018squeeze}—to dynamically produce scaling vectors conditioned on each sample's feature map.
    }
    \label{fig:scaleu-all}
    \end{figure*}
}

\def\figIterativeImgGen#1{
    \captionsetup[sub]{font=small}
    \begin{figure*}[#1]
      \centering
      \includegraphics[width=0.95\linewidth]{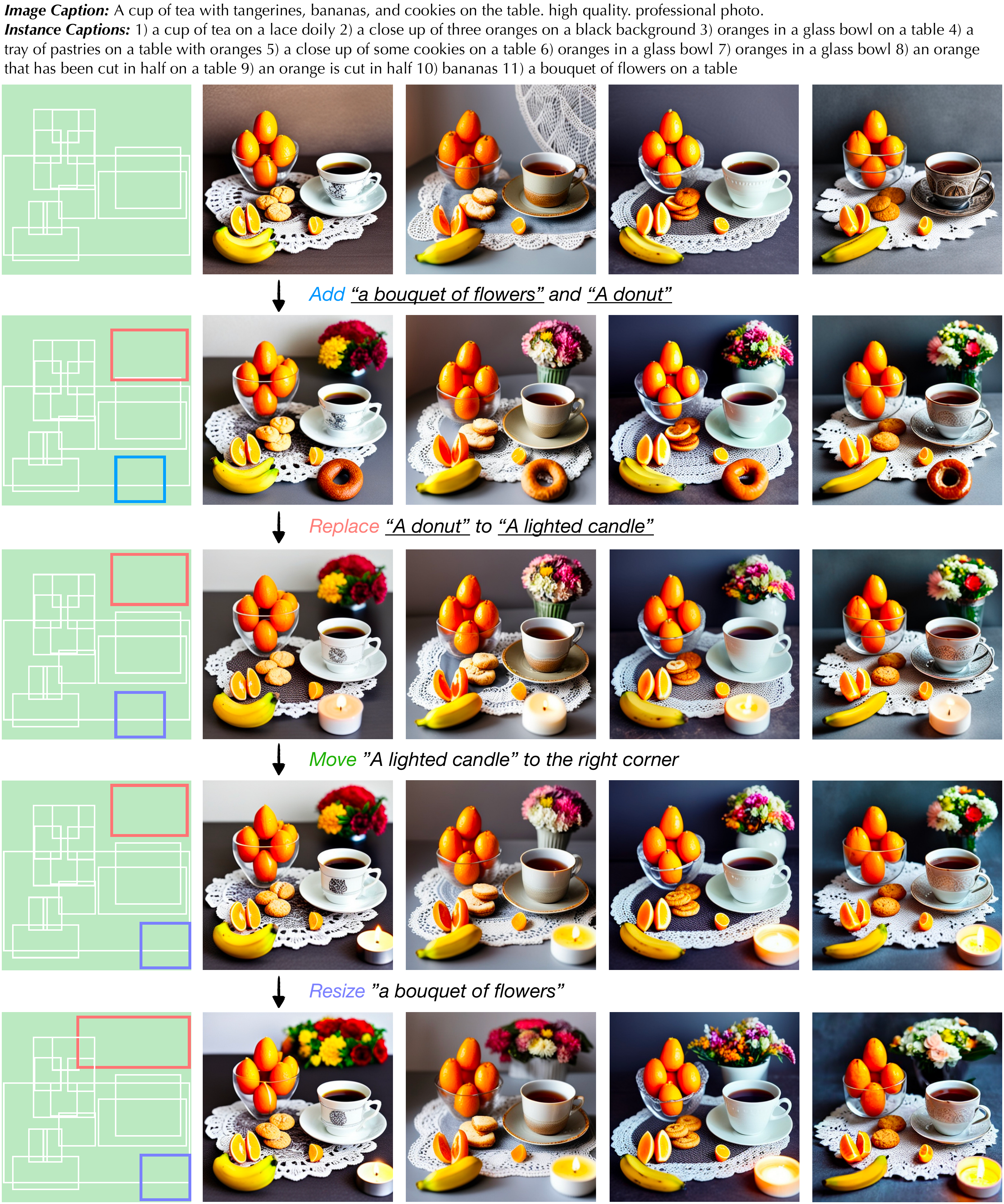}
      \caption{Iterative Image Generation. 
      With minimal changes to pre-generated instances and the overall scene, users can selectively introduce new instances (as seen in row two, where ``a bouquet of flowers'' and ``a donut'' are added to the images from row one), substitute one instance for another (in row three, ``a donut'' is replaced with ``a lighted candle''), reposition an instance (in row four, ``a lighted candle'' is moved to the bottom right corner), or adjust the size of an instance (in row five, the size of ``a bouquet of flowers'' is increased).
      }
      \label{fig:iterative_img_gen}
    \end{figure*}
}

\section{Preliminary}
\label{sec:preliminaries}

\noindent \textbf{Diffusion Models}~\cite{ho2020denoising,sohl2015deep,song2020score} learn the process of text-to-image generation through iterative denoising steps initiated from an initial random noise map, denoted as $z_{T}$.
Latent diffusion models (LDMs)~\cite{rombach2021highresolution} perform the diffusion process in the latent space of a Variational AutoEncoder~\cite{DBLP:journals/corr/KingmaW13}, for computational efficiency, and encode the textual inputs as feature vectors from pretrained language models~\cite{radford2021learning,raffel2020exploring}.

Specifically, starting from a noised latent vector $\mathbf{z}_t$ at the time step $t$, a denoising autoencoder~\cite{rombach2021highresolution,ronneberger2015u}, denoted as $\epsilon_{\theta}$, is trained to predict the noise $\epsilon$ that is added to the latent vector $\mathbf{z}$, conditioned on the text prompt $\mathbf{c}$. The training objective is defined as:
\begin{equation}
    \mathcal{L} = \mathbb{E}_{\mathbf{z}\sim \mathcal{E}(\mathbf{x}),\epsilon \sim \mathcal{N}(0,1),t}\left[||\epsilon - \epsilon_{\theta}(\mathbf{z}_t,t,\tau(\mathbf{c}))||^2_2\right],
\end{equation}
where $t$ is uniformly sampled from the set of time steps $\{1,...,T\}$.
$\tau$ pre-process the text prompt $\mathbf{c}$ into text tokens $\tau(\mathbf{c})$, utilizing the pretrained CLIP text model~\cite{radford2021learning}.

During inference, a latent vector $\mathbf{z}_T$, sampled from a standard normal distribution $\mathcal{N}(0,1)$ is iterative denoised using DDIM~\cite{song2020denoising} to obtain $z_0$.
Finally, the latent vector $\mathbf{z}_0$ is input into the decoder of VAE to generate an image $\widetilde{\mathbf{x}}$.

\section{Ablation Study}
In addition to the ablation study we presented in~\cref{sec:exp-abl}, in this section, we also offer additional ablations focusing on the hyper-parameters of \unifusion modules, design variations for \scaleu, the impact of model inference with hybrid inputs, among other aspects.

\tabAblationsUniFusion{!h}

\noindent \textbf{Design choices for \unifusion.}
We first analyze the impact of frequency bandwidths when projecting location conditions into a higher-dimensional feature space with Fourier Transform, as depicted in~\cref{tab:ablate_f_band_sup}. 
The Fourier transform process empowers a multilayer perceptron (MLP) to grasp high-frequency functions in low-dimensional problem domains~\cite{tancik2020fourfeat}.
We apply the Fourier mapping to the 2D point coordinates associated with each location to convert them into an embedding.
The embedding enables MLPs to better learn a high-frequency function for the coordinates.
Notably, expanding the frequency bandwidth tends to improve the performance, but a plateau is reached once the bandwidth exceeds 16.
The influence of the dimensionality ($N$) of the MLP layer within \unifusion is assessed in \cref{tab:ablate_unifusion_dim_sup}. We find that a dimension of 3072 emerges as the optimal balance between model efficacy and its size. Increasing the MLP layers dimensions from 3072 to 4096 does not yield further improvements in performance. Therefore, we select $N\!=\!3072$ by default.

\noindent \textbf{\textit{Can we use one single token for all location conditions?}}
Actually, we can still achieve reasonable performance using a unified tokenization function that results in a single token for all forms of location inputs, as demonstrated in ~\cref{tab:abl-components}.
However, having multiple tokens ($M$ tokens) for different input types ($M$ types) leads to optimal performance. This is because these four types of layout conditions necessitate distinct approaches to ensuring that the model respects the layout condition appropriately. Specifically, the model needs to disseminate grounding information to adjacent visual tokens when using point and scribble inputs.
In contrast, bounding-box and mask conditions require the model to confine the grounding information injection within the specified box or mask.

\noindent \textbf{\textit{Why not employ masks as extra channels, as seen in GLIGEN~\cite{li2023gligen} and ControlNet~\cite{zhang2023adding}?}}
In these approaches, the semantic segmentation masks (do not discriminate instances in the same class) are resized to a smaller resolution of $64\times64$ features. Nonetheless, our observations indicate that when the occlusion ratio between instances is high, particularly in cases where overlapping instances carry similar semantic information, the model's performance is compromised a lot. 
Additionally, the model encounters difficulties when generating high-quality results for very small objects. Therefore, we convert all masks into point-based inputs.
However, it is possible that adding segmentation masks as additional input could further improve our model's performance, we leave it for future research.

\tabAblationsScaleU{!h}
\figScaleUDiagram{!tb}

\noindent \textbf{Design choices for ScaleU} are depicted in~\cref{fig:scaleu-all}. Beyond the standard ScaleU block described in~\cref{sec:method-scaleu}, which re-calibrates both main and skip-connected features before their concatenation in the UNet model, we explored an alternative design, SE-ScaleU (~\cref{fig:scaleu-all}c). This variant employs an MLP layer, similar to the Squeeze-and-Excitation module~\cite{hu2018squeeze}, for dynamically generating scaling vectors based on each sample's feature map. 
However, as demonstrated in~\cref{tab:ablate_scaleu}, while SE-ScaleU offers performance on par with the light-weight ScaleU block, it requires additional parameters in the MLP layers. Consequently, we default to using ScaleU. 

\tabMISDesign{!h}
\noindent \textbf{Design choices for \mis.} There are two design strategies for \mis: crop-and-paste and instance latents averaging, with the latter being our paper's default approach. The crop-and-paste \mis involves:
1) Running separate denoising operations for each of the $n$ instances over $M$ steps to obtain instance latents $L_I$.
2) Cropping instance latents $\{L_I^1,\cdots,L_I^n\}$ as per location conditions and pasting these cropped, denoised latents onto the global latent $L_G$, derived from all instance tokens and text prompts, at their respective locations.
3) Continuing the denoising process on the combined latent from step (2) using all instance tokens, instance text prompts, and the global image prompt.
This process largely mirrors our default latent averaging \mis, except for step (2)'s latent merging method.

While crop-and-paste \mis matches or slightly surpasses the performance of our default averaging approach on some testing cases, it has its limitations: 
1) In step (2) of the crop-and-paste \mis, the model needs to crop instance latents according to the bounding box or mask provided, limiting its application to bounding boxes, and instance masks. For point inputs and \scribble{}s, the model has to conjecture the size/shape of the instance. 
2) The presence of overlapping instances presents a challenge. The model can only preserve latents from a single instance in these regions, resulting in blurred and diminished-quality pixels in areas of instance overlap.

\figDemoSteps{!tbh}

\tabHybridInputs{!bth}
\noindent \textbf{Multiple location formats at inference} are analyzed in~\cref{tab:abl-hybrid}. It is observed that having more location conditions provides the best performance and more precise control on the instance location.
This results in significant performance improvements, particularly for instance masks (9.9\% AP$^\text{mask}$) and \scribble (16.3\% PiM).
Note that many of the other location formats can be automatically derived: 
For image generation conditioned on instance masks, since both the box and the central point can be inferred from the mask, our model enjoys this performance improvement without imposing extra demands on users; 
Likewise, for boxes, the performance gains achieved by incorporating a point as the instance location condition can be obtained without any additional user inputs.
These derived location formats improve location conditioning without additional user inputs.

\noindent \textbf{Impact of \unifusion module}. 
\cref{fig:unifuion_steps} illustrates that as the UniFusion module is applied over a increasing percentage of timesteps (ranging from 10\% to 75\%), the model’s adherence to the instance conditions progressively improves.  
For instance, the sunflower in the top left corner is generated only when the UniFusion module is active for 75\% of the total timesteps. Similarly, the sunflower in the bottom right corner manifests after the module has been active for 25\% of the timesteps. Additionally, the model's ability to accurately adhere to the teddy bear's location condition is enhanced as \unifusion is utilized for more extended timesteps.

\figIterativeImgGen{!tb}
\figDemoPose{!t}

\figDemoPointScribble{!t}
\figMoreDemos{!t}

\section{Model Training}
\noindent \textbf{Model training.}
We follow the same setup as GLIGEN~\cite{li2023gligen} and initialize our model with a pretrained \textToI model whose layers are kept frozen.
We add the learnable parameters for instance conditioning and train the model with a batch size of 512 for 100K steps. 
We use the Adam optimizer~\cite{kingma2014adam} with a learning rate that is warmed up to $0.0001$ after $5000$ iterations.
We learn the model with exponential moving average (EMA) on model parameters with a decay factor of 0.99 and use the EMA model during the inference time.
In addition, we have a 10\% probability to set all four location inputs as null tokens to support classifier-free guidance, following the approach proposed in~\cite{ho2022classifier}. 
Additionally, for the various location condition tokens, including masks, bounding boxes, points, and \scribble{}s, each has a 10\% dropout rate.
We use 64 Nvidia A100 GPUs to train the model. 

\textbf{}

\section{Applications and Qualitative Results}

\noindent \textbf{Iterative Image Generation.} 
InstanceDiffusion's capability for precise instance control allows \ours to excel in multi-round image generation, leveraging this feature. 
\ours enables users to strategically place objects in specific locations while maintaining the consistency of previously generated objects and the overall scene. We outline the process of our iterative image generation in the following three steps:

\begin{itemize}[leftmargin=*,nosep]
    \item 1) Initially, generate images using the global image caption, all instance captions with their respective location conditions, and random noise.
    \item 2) Users have the option to introduce new instances by supplying additional instance conditions, including text prompts and locations. They can also modify existing instances by altering their descriptions or locations.
    \item 3) Employ the revised set of instance conditions, the global prompt, and the same random noise as in step 1 to create a new image.
\end{itemize}
Steps 2 and 3 can be repeated for multiple rounds until the desired outcome is achieved. 

In addition to the visuals we have shown in the main paper, we provide more qualitative results on iterative image generation in ~\cref{fig:iterative_img_gen}.
With minimal changes to pre-generated instances and the overall scene, users can selectively introduce new instances (as seen in row two, where “a bouquet of flowers” and “a donut” are added to the images from row one), substitute one instance for another (in row three, “a donut” is replaced with “a lighted candle”), reposition an instance (in row four, “a lighted candle” is moved to the bottom right corner), or adjust the size of an instance (in row five, the size of “a bouquet of flowers” is increased).

\noindent \textbf{Hierarchical location conditioning in image composition.} 
Our findings, illustrated in ~\cref{fig:demo_pose}, reveal that incorporating hierarchical location conditionings - specifically, the locations and sizes of parts and subparts of an instance - as model inputs subtly alters the overall pose of an object (right, left, front). This demonstrates the effective use of spatial hierarchy in visual design. 
We hope that this capability could inspire more future research and applications in fine-grained control in image generation. 

\noindent \textbf{More demo} results for InstanceDiffusion's image generation are shown in \cref{fig:demo_supp_ex1,fig:demo_supp_ex2}.

\clearpage
\cleardoublepage


\end{document}